\def\year{2020}\relax
\documentclass[letterpaper]{article}

\usepackage{aaai20}
\usepackage{times}
\usepackage{helvet}
\usepackage{courier}
\usepackage[hyphens]{url}
\usepackage{graphicx}
\usepackage{graphicx}
\usepackage[framemethod=tikz]{mdframed}
\usepackage{amssymb}
\usepackage{amsmath}
\usepackage[ruled,linesnumbered]{algorithm2e}
\usepackage{amsfonts}
\usepackage{amsthm}
\usepackage{bm}
\usepackage{bbm}
\usepackage{subcaption}
\usepackage{multirow}
\usepackage{adjustbox}
\usepackage{array, booktabs}

\def\eqref#1{equation~\ref{#1}}

\def\1{\bm{1}}

\def\rvx{{\mathbf{x}}}

\def\vzero{{\bm{0}}}

\def\vtheta{{\bm{\theta}}}

\def\vs{{\bm{s}}}

\def\mI{{\bm{I}}}

\DeclareMathAlphabet{\mathsfit}{\encodingdefault}{\sfdefault}{m}{sl}
\SetMathAlphabet{\mathsfit}{bold}{\encodingdefault}{\sfdefault}{bx}{n}

\newcommand{\E}{\mathbb{E}}

\newcommand{\norm}[1]{\left\lVert#1\right\rVert}

\newcommand{\RN}{\mathbb{R}}

\newcommand{\AC}{\mathcal{A}}

\newcommand{\citet}[1]{\citeauthor{#1} \shortcite{#1}}
\newcommand{\citep}{\cite}

\title{Proximal Distilled Evolutionary Reinforcement Learning}
\author{\Large \textbf{Cristian Bodnar, Ben Day, Pietro Li\'{o}}\\ %
Department of Computer Science \& Technology\\
University of Cambridge\\
Cambridge, United Kingdom\\
cb2015@cam.ac.uk }
\begin{document}
\maketitle

\begin{abstract}
Reinforcement Learning (RL) has achieved impressive performance in many complex environments due to the integration with Deep Neural Networks (DNNs). At the same time, Genetic Algorithms (GAs), often seen as a competing approach to RL, had limited success in scaling up to the DNNs required to solve challenging tasks. Contrary to this dichotomic view, in the physical world, evolution and learning are complementary processes that continuously interact. The recently proposed Evolutionary Reinforcement Learning (ERL) framework has demonstrated mutual benefits to performance when combining the two methods. However, ERL has not fully addressed the scalability problem of GAs. In this paper, we show that this problem is rooted in an unfortunate combination of a simple genetic encoding for DNNs and the use of traditional biologically-inspired variation operators. When applied to these encodings, the standard operators are destructive and cause catastrophic forgetting of the traits the networks acquired. We propose a novel algorithm called Proximal Distilled Evolutionary Reinforcement Learning (PDERL) that is characterised by a hierarchical integration between evolution and learning. The main innovation of PDERL is the use of learning-based variation operators that compensate for the simplicity of the genetic representation. Unlike traditional operators, our proposals meet the functional requirements of variation operators when applied on directly-encoded DNNs. We evaluate PDERL in five robot locomotion settings from the OpenAI gym. Our method outperforms ERL, as well as two state-of-the-art RL algorithms, PPO and TD3, in all tested environments.
\end{abstract}

\section{Introduction}

\noindent The field of Reinforcement Learning (RL) has recently achieved great success by producing artificial agents that can master the game of Go \citep{Silver2017MasteringTG}, play Atari games \citep{Mnih2015HumanlevelCT} or control robots to perform complex tasks such as grasping objects \citep{Andrychowicz2017HindsightER} or running \citep{Lillicrap2015ContinuousCW}. Most of this success is caused by the combination of RL with Deep Learning \citep{Goodfellow-et-al-2016}, generically called Deep Reinforcement Learning (DRL). 

At the same time, Genetic Algorithms (GAs), usually seen as a competing approach to RL, have achieved limited success in evolving DNN-based control policies for complex environments. Though previous work has shown GAs to be competitive with other DRL algorithms in discrete environments \citep{Such2017DeepNG}, they are still significantly less sample efficient than a simple method like Deep Q-Learning \citep{Mnih2015HumanlevelCT}. Moreover, in complex robotic environments with large continuous state and action spaces, where environment interactions are costly, their sample inefficiency is even more acute \citep{Khadka2018EvolutionGuidedPG}, \citep{Such2017DeepNG}. 

However, in the physical world, evolution and learning interact in subtle ways. Perhaps, the most famous product of this interaction is the Baldwin effect \citep{simpson_baldwin}, which explains how the genotype can assimilate learnt behaviours over the course of many generations. A more spectacular by-product of this interplay, which has received more attention in recent years, is the epigenetic inheritance of learnt traits \citep{mice_fear_nature}.

Despite these exciting intricacies of learning and evolution, the two have almost always received separate treatment in the field of AI. Though they have been analysed together in computational simulations multiple times \citep{Hinton1987HowLC}, \citep{Ackley+Littman:1992}, \citep{Suzuki2004InteractionsBL}, they have rarely been combined to produce novel algorithms with direct applicability. This is surprising given that nature has always been a great source of inspiration for AI \citep{Floreano:2008:BAI:1457317}.

For the first time, \citet{Khadka2018EvolutionGuidedPG} have recently demonstrated on robot locomotion tasks the practical benefits of merging the two approaches in their Evolutionary Reinforcement Learning (ERL) framework. ERL uses an RL-based agent alongside a genetically evolved population, with a transfer of information between the two. However, ERL has not fully addressed the scalability problem of GAs. While the gradient information from the RL agent can significantly speed up the evolutionary search, the population of ERL is evolved using traditional variation operators. Paired with directly encoded DNNs, which is the most common genetic representation in use, we show that these operators are destructive.

This paper brings the following contributions:
\begin{itemize}
\item Demonstrates the negative side-effects in RL of the traditional genetic operators when applied to directly encoded DNNs.
\item Proposes two novel genetic operators based on backpropagation. These operators do not cause catastrophic forgetting in combination with simple DNN representations. 
\item Integrates these operators as part of a novel framework called Proximal Distilled Evolutionary Reinforcement Learning (PDERL) that uses a hierarchy of interactions between evolution and learning.
\item Shows that PDERL outperforms ERL, PPO \citep{Schulman2017ProximalPO} and TD3 \citep{Fujimoto2018AddressingFA} in five robot locomotion environments from the OpenAI gym \citep{gym}. 
\end{itemize}

\section{Background}

This section introduces the Evolutionary Reinforcement Learning (ERL) algorithm and the genetic operators it uses.

\subsection{Evolutionary Reinforcement Learning}

The proposed methods build upon the ERL framework introduced by \citet{Khadka2018EvolutionGuidedPG}. In this framework, a population of policies is evolved using GAs. The fitness of the policies in the population is based on the cumulative total reward obtained over a given number of evaluation rounds. Alongside the population, an actor-critic agent based on DDPG \citep{Lillicrap2015ContinuousCW} is trained via RL. The RL agent and the population synchronise periodically to establish a bidirectional transfer of information.   

The first type of synchronisation in ERL, from the RL agent to the genetic population, is meant to speed up the evolutionary search process. This synchronisation step clones the actor of the RL agent into the population every few generations to transfer the policy gradient information. The synchronisation period, $\omega$, is a hyperparameter that controls the rate of information flowing from the RL agent to the population.

The second type of synchronisation consists of a reverse information flow coming from the population to the RL agent. The actors in the population collect experiences from which the RL agent can learn off-policy. All the transitions coming from rollouts in the population are added to the replay buffer of the DDPG agent. The population experiences can be seen as being generated by evolution-guided parameter space noise \citep{Plappert2018ParameterSN}.

\subsection{Genetic encoding and variation operators}

The policies in the ERL population are represented by neural networks with a direct encoding. In this common genetic representation, the weights of a network are recorded as a list of real numbers. The ordering of the list is arbitrary but consistent across the population. As we will show, applying the usual biologically inspired variation operators on this representation can produce destructive behaviour modifications. 

In the physical world, mutations and crossovers rarely have catastrophic phenotypic effects because the phenotype is protected by the complex layers of physical, biological and chemical processes that translate the DNA. In a direct genetic encoding, the protective layers of translation are absent because the representation is so simple and immediate. As such, the biologically inspired variation operators commonly found in the literature, including ERL, do not have the desired functionality when paired with a direct encoding. Ideally, crossovers should combine the best behaviours of the two parents. At the same time, mutations should produce only a slight variation in the behaviour of the parent, ensuring that the offspring inherits it to a significant extent. However, because DNNs are sensitive to small modifications of the weights (the genes in a direct encoding), these operators typically cause catastrophic forgetting of the parental behaviours.

ERL evolves the population using two variation operators commonly used for list-based representations: $n$-point crossovers and Gaussian mutations \citep[p. 49-79]{Eiben:2015:IEC:2810085}. $n$-point crossovers produce an offspring policy by randomly exchanging segments of the lists of weights belonging to the two parents, where $n$ endpoints determine the segments. ERL uses a version of the operator where the unit-segments are rows of the dense layer matrices, ensuring that an offspring receives nodes as they appear in the parents rather than splicing the weights of nodes together. The resulting child policy matrices contain a mix of rows (nodes) coming from the matrices (layers) of both parents. This is intended to produce functional consistency across generations. 

However, the lack of an inherent node ordering in DNNs means that hidden representations need not be consistent over the population and as such the input to a node may not be consistent from parent to offspring, creating the possibility for destructive interference. This can cause the child policy to diverge from that of the parents, as we will demonstrate. Similarly, the damaging effects of adding Gaussian noise to the parameters of a DNN have been discussed at great length by \citet{Lehman2018SafeMF}. A common approach to containing these issues, employed by ERL, is to mutate only a fraction of the weights. Nevertheless, these mutations are still destructive. Furthermore, evolving only a small number of weights can slow down the evolutionary search for better policies. 

\section{Method}

This section introduces our proposed learning-based genetic operators and describes how they are integrated with ERL.  

\subsection{The genetic memory}
\label{sec:genetic_memory}

A significant problem of the population from ERL is that it does not directly exploit the individual experiences collected by the actors in the population. The population only benefits indirectly, through the RL agent, which uses them to learn and improve. The individual experiences of the agents are an essential aspect of the new operators we introduce in the next sections and, therefore, the agents also need a place to store them. 

The first modification we make to ERL is to equip the members of the population, and the RL agent, with a small personal replay buffer containing their most recent experiences, at the expense of a marginally increased memory footprint. Depending on its capacity $\kappa$, the buffer can also include experiences of their ancestors. Because the transitions in the buffer can span over multiple generations, we refer to this personal replay buffer of each agent as the \textit{genetic memory}. When the policies interact with the environment, they not only store their experiences in DDPG's replay buffer as in ERL but also in their \textit{genetic memory}. 

The ancestral experiences in the genetic memory are introduced through the variation operators. A mutated child policy inherits the genetic memory of the parent entirely. During crossover, the buffer is only partially inherited. The crossover offspring fills its buffer with the most recent half of transitions coming from each of the two parents' genetic memories.

\subsection{Q-filtered distillation crossovers}
\label{sec:new_crossover}

In this section, we propose a \textit{$Q$-filtered behaviour distillation crossover} that selectively merges the behaviour of two parent policies into a child policy. Unlike $n$-point crossovers, this operator acts in the phenotype space, and not in parameter space. 

\begin{figure}[ht]
     \centering
     \includegraphics[width=1.0\columnwidth]{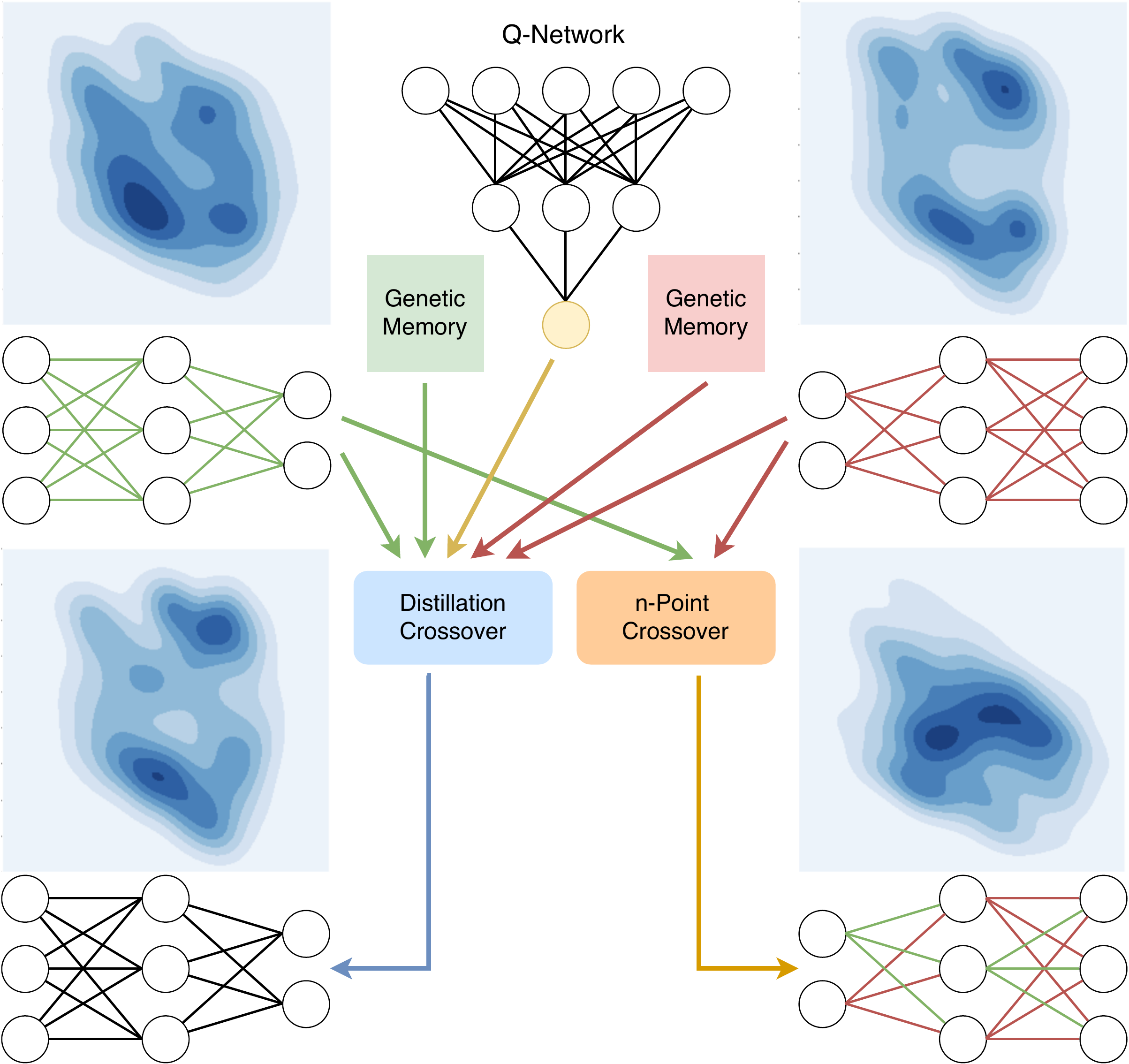}
     \caption{$Q$-filtered distillation crossover compared to $n$-point crossover. The contour plots represent the state visitation distributions of the agents (i.e. the average amount of time the agent spends in each state) for the first two state dimensions of an environment. The plot is generated by fitting a Gaussian kernel density model over the states collected over many episodes. These distributions show how $Q$-filtered distillation crossover selectively merges the behaviours of the two parents by inheriting from the shapes of both parent distributions. In contrast, the modes of the state visitation distribution obtained by the traditional crossover are mostly disjoint from the modes of the parent distributions. }
     \label{fig:distil_crossover}
\end{figure}

For a pair of parent actors from the population, the crossover operation works as follows. A new agent with an initially empty associated genetic memory is created. The genetic memory is filled in equal proportions with the latest transitions coming from the genetic memories of the two parents. The child agent is then trained via a form of Imitation Learning \citep{Osaetal18} to selectively imitate the actions the parents would take in the states from the newly created genetic memory. Equivalently, this process can be seen as a more general type of policy distillation \citep{Rusu2016PolicyD} process since it aims to ``distil'' the behaviour of the parents into the child policy. 

\begin{algorithm}[h]
\SetAlgoLined
\SetKwInOut{Input}{Input}\SetKwInOut{Output}{Output}

\Input{Parent policies $\mu_x, \mu_y$ with memory $R_x, R_y$} 
\Output{Child policy $\mu_z$ with memory empty $R_z$} 
 Add latest $\frac{\kappa}{2}$ transitions from $R_x$ to $R_z$ \\
 Add latest $\frac{\kappa}{2}$ transitions from $R_y$ to $R_z$ \\
 Shuffle the transitions in $R_z$\\
 Randomly initialise the weights $\vtheta_z$ of $\mu_z$ with the weights of one of the parents \\ 
 
 \For{$e\leftarrow 1$ \KwTo epochs}{
    \For{$i\leftarrow 1$ \KwTo $\kappa / N_C$}{
        Sample state batch of size $N_C$ from $R_z$ \\
        Optimise $\vtheta_z$ to minimise $L(C)$ using SGD \\
    }
 }
 \caption{Distillation Crossover}
\end{algorithm}

Unlike the conventional policy distillation proposed by \citet{Rusu2016PolicyD}, two parent networks are involved, not one. This introduces the problem of divergent behaviours. The two parent policies can take radically different actions in identical or similar states. The problem is how the child policy should decide whom to imitate in each state. The key observation of the proposed method is that the critic of the RL agent already knows the values of certain states and actions. Therefore, it can be used to select which actions should be followed in a principled and globally consistent manner. We propose the following $Q$-filtered behaviour cloning loss to train the child policy:  
\begin{equation*}
\begin{split}
    L(C) &= \sum_i^{N_C} \norm{\mu_z(s_i) - \mu_x(s_i)}^2 \mathbb{I}_{Q(s_i, \mu_x(s_i)) > Q(s_i, \mu_y(s_i))} \\
    &+ \sum_j^{N_C} \norm{\mu_z(s_j) - \mu_y(s_j)}^2 \mathbb{I}_{Q(s_j, \mu_y(s_j)) > Q(s_j, \mu_x(s_j))} \\
    &+ \frac{1}{N_{C}}\sum_k^{N_C} \norm{\mu_z(s_k)}^2 \text{,}
\end{split}
\raisetag{2\normalbaselineskip}
\end{equation*}
where the sum is taken over a batch of size $N_{C}$ sampled from the genetic memories of the two parent agents. $\mu_x$ and $\mu_y$ represent the deterministic parent policies, while $\mu_z$ is the deterministic policy of the child agent.

The indicator function $\mathbb{I}$ uses the $Q$-Network of the RL agent to decide which parent takes the best action in each state. The child policy is trained to imitate those actions by minimising the first two terms. The final term is an $L_2$ regularisation that prevents the outputs from saturating the hyperbolic tangent activation. Figure \ref{fig:distil_crossover} contains a diagram comparing this new crossover with the ERL $n$-point crossover. We refer to ERL with the distillation crossover as Distilled Evolutionary Reinforcement Learning (DERL).

We note that while this operator is indeed more computationally intensive, a small number of training epochs over the relatively small genetic memory suffices. Additionally, we expect a distributed implementation of our method to compensate for the incurred wall clock time penalties. We leave this endeavour for future work. 


\subsection{Parent selection mechanism}

An interesting question is how parents should be selected for this crossover. A general approach is to define a mating score function $m: \Pi \times \Pi \to \RN$ that takes as input two policies and provides a score. The pairs with higher scores are more likely to be selected. Similarly to \citet{Gangwani2018PolicyOB}, we distinguish two ways of computing the score: greedy and distance-based. 

\textbf{Greedy} The score $m(\mu_x, \mu_y) = f(\mu_x) +  f(\mu_y)$ can be greedily determined by the sum of the fitness of the two parents. This type of selection generally increases the stability of the population and makes it unlikely that good individuals are not selected.

\textbf{Distance based} The score $m(\mu_x, \mu_y) = d_\Pi(\mu_x, \mu_y)$ can be computed using a distance metric in the space of all possible policies. ``Different'' policies are more likely to be selected for mating. The exact notion of ``different'' depends on the precise form of the distance metric $d_\Pi$. Here, we propose a distance metric in the behaviour space of the two policies that takes the form:
\begin{equation*}
\begin{split}
    d_\Pi(\mu_x, \mu_y) &= \E_{\rvx \sim \rho_x}[\norm{\mu_x(\rvx) - \mu_y(\rvx)}^2] \\ 
    &+ \E_{\rvx \sim \rho_y}[\norm{\mu_x(\rvx) - \mu_y(\rvx)}^2] \text{,}
\end{split}
\end{equation*}
where $\rho_x$ and $\rho_y$ are the state-visitation distributions of the two agents.

This distance metric measures the expected difference in the actions taken by the two parent policies over states coming from a mixture of their state visitation distributions. This expectation is in practice stochastically approximated by sampling a large batch from the genetic memories of the two agents. This strategy biases the introduction of novel behaviours into the population at the expense of stability as the probability that fit individuals are not selected is increased.


\subsection{Proximal mutations}
\label{sec:new_mutation}

As showed by \citet{Lehman2018SafeMF}, Gaussian mutations can have catastrophic consequences on the behaviour of an agent. In fact, the stability of the policy update is a problem even for gradient descent approaches, where an inappropriate step size can have unpredictable consequences in the performance landscape. Methods like PPO \citep{Schulman2017ProximalPO} are remarkably stable by minimising an auxiliary KL divergence term that keeps the behaviour of the new policy close to the old one.

\begin{figure}[ht]
     \centering
     \includegraphics[width=1.0\columnwidth]{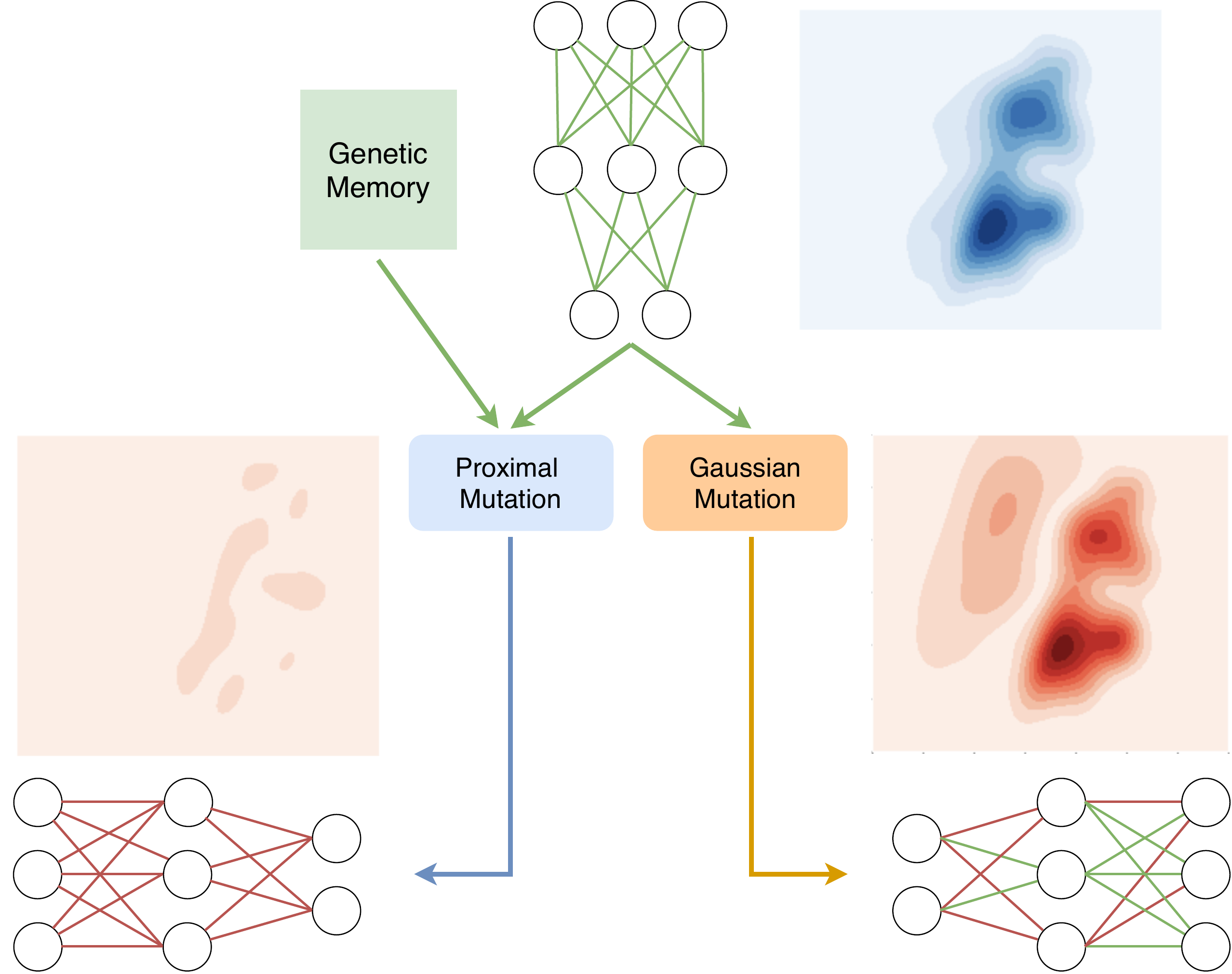}
     \caption{Proximal mutations compared to Gaussian mutations. The blue contour plot shows the state visitation distribution of the parent policy. The red contour plots show the difference between the distribution of the children and that of the parent. The difference plots are generated by taking the normalised difference between the parent and child probability densities.  The behaviour of the policy obtained by proximal mutation is a small perturbative adjustment to the parent behaviour. In contrast, the traditional mutation produces a divergent behaviour, even though it modifies only a fraction of the weights (shown in red). }
     \label{fig:proximal_mutation}
\end{figure}
\begin{algorithm}
\SetAlgoLined
\SetKwInOut{Input}{Input}\SetKwInOut{Output}{Output}

\Input{Parent policy $\mu_x$ with memory $R_x$} 
\Output{Child policy $\mu_y$ with memory $R_y$} 
 Initialise $R_y \leftarrow R_x$ and $\mu_y \leftarrow \mu_x$ \\ 
 Sample state batch of size $N_M$ from $R_x$ \\ 
 Compute $\vs$ on the batch samples $s_i$ as in Equation \ref{eq:proximal_mut}. \\
 Mutate $\vtheta_y \leftarrow \vtheta_y + \frac{\rvx}{\vs}, \rvx \sim \mathcal{N}(\vzero, \sigma \mI)$
 \caption{Proximal Mutation}
\end{algorithm}

Based on these motivations, we integrate the safe mutation operator SM-G-SUM that has been proposed by \citet{Lehman2018SafeMF} with the genetic memory of the population. This operator uses the gradient of each dimension of the output action over a batch of $N_M$ transitions from the genetic memory to compute the sensitivity $\vs$ of the actions to weight perturbations: 
\begin{equation}
\label{eq:proximal_mut}
    \vs = \sqrt{\sum_k^{|\AC|} \Big(\sum_i^{N_M} \nabla_\vtheta \mu_\vtheta(s_i)_k\Big)^2} 
\end{equation}
The sensitivity is then used to scale the Gaussian perturbation of each weight accordingly by $\vtheta \leftarrow \vtheta + \frac{\rvx}{\vs}$, with $\rvx \sim \mathcal{N}(\vzero, \sigma \mI)$, where $\sigma$ is a mutation magnitude hyperparameter. The resulting operator produces child policies that are in the \textit{proximity} of their parent's behaviour. Therefore, we refer to this operator as a \textit{proximal mutation} (Figure \ref{fig:proximal_mutation}), and the version of ERL using it as Proximal Evolutionary Reinforcement Learning (PERL). 

While the proximal mutations do not explicitly use learning, they rely on the capacity of the policies to learn, or in other words, to be differentiable. Without this property, these behaviour sensitivities to the parameter perturbations cannot be computed analytically.

\subsection{Integration}

The full benefits of the newly introduced operators are realised when they are used together. The $Q$-filtered distillation crossover increases the stability of the population and drives the agents towards regions of the state-action space with higher $Q$-values. The proximal mutations improve the exploration of the population and its ability to discover better policies. As will be seen in the evaluation section, the operators complement each other. We refer to their dual integration with ERL as Proximal Distilled Evolutionary Reinforcement Learning (PDERL).

\begin{figure}[ht]
    \centering
    \includegraphics[width=0.8\columnwidth]{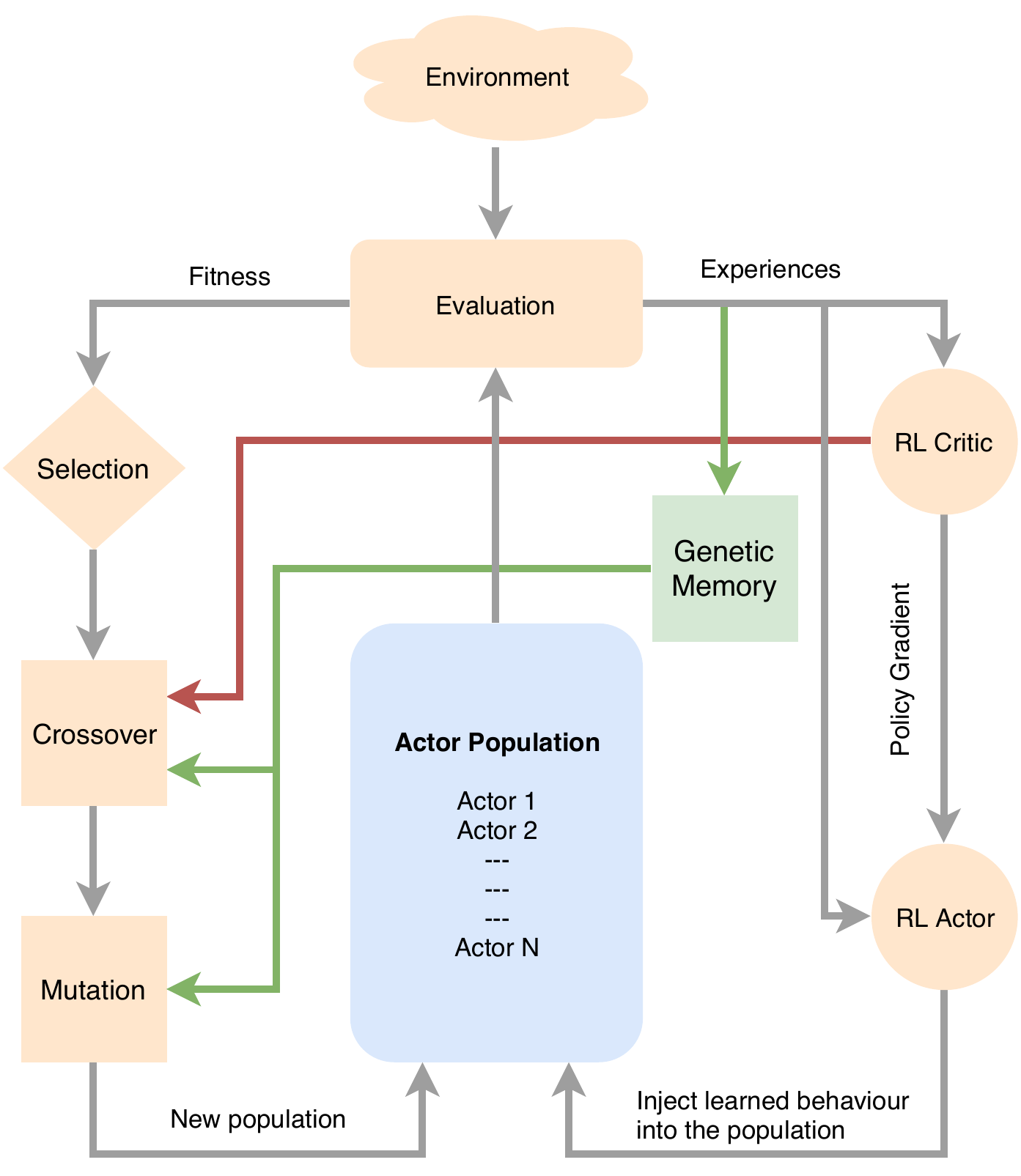}
    \caption[Proximal Distilled Evolutionary Reinforcement Learning]{A high-level view of PDERL. The new components and interactions are drawn in green and red. In PDERL, there is a higher flow of information from the individual experiences and learning (right) to the population (left) than in ERL.}
    \label{fig:pderl}
\end{figure}

Ultimately, PDERL contains a hierarchy of interactions between learning and evolution. A high-level interaction is realised through the information exchange between the population and the RL agent. The newly introduced operators add a lower layer of interaction, at the level of the genetic operators. A diagram of PDERL is given by Figure \ref{fig:pderl}.

\section{Evaluation}

This section evaluates the performance of the proposed methods, and also takes a closer look at the behaviour of the proposed operators.

\subsection{Experimental setup}

The architecture of the policy and critic networks is identical to ERL. Those hyperparameters that are shared with ERL have the same values as those reported by \citet{Khadka2018EvolutionGuidedPG}, with a few exceptions. For Walker2D, the synchronisation rate $\omega$ was decreased from $10$ to $1$ to allow a higher information flow from the RL agent to the population. In the same environment, the number of evaluations $\xi$ was increased from $3$ to $5$ because of the high total reward variance across episodes. Finally, the fraction of elites in the Hopper and Ant environments was reduced from $0.3$ to $0.2$. Generally, a higher number of elites increases the stability of the population, but the stability gained through the new operators makes higher values of this parameter unnecessary.

For the PDERL specific hyperparameters, we performed little tuning due to the limited computational resources. In what follows we report the chosen values alongside the values that were considered. The crossover and mutation batch sizes are $N_{C} = 128$ and $N_{M} = 256$ (searched over $64, 128, 256$). The genetic memory has a capacity of $\kappa = 8k$ transitions ($2k, 4k, 8k, 10k$). The learning rate for the distillation crossover is $10^{-3}$ ($10^{-2}, 10^{-3}, 10^{-4}, 10^{-5}$), and the child policy is trained for $12$ epochs ($4, 8, 12, 16$) . All the training procedures use the Adam optimiser. Greedy parent selection is used unless otherwise indicated. As in ERL, the population is formed of $k=10$ actors. 

When reporting the results, we use the official implementations for  ERL\footnote{\url{https://github.com/ShawK91/erl_paper_nips18}} and TD3\footnote{\url{https://github.com/sfujim/TD3}}, and the OpenAI Baselines\footnote{\url{https://github.com/openai/baselines/}} implementation for PPO. Our code is publicly available at \url{https://github.com/crisbodnar/pderl}.

\subsection{Performance evaluation}
\label{sec:peformance_evaluation}

\begin{figure*}[!ht]
    \centering
        \begin{subfigure}[t]{0.30\textwidth}
         \centering
         \includegraphics[width=\columnwidth]{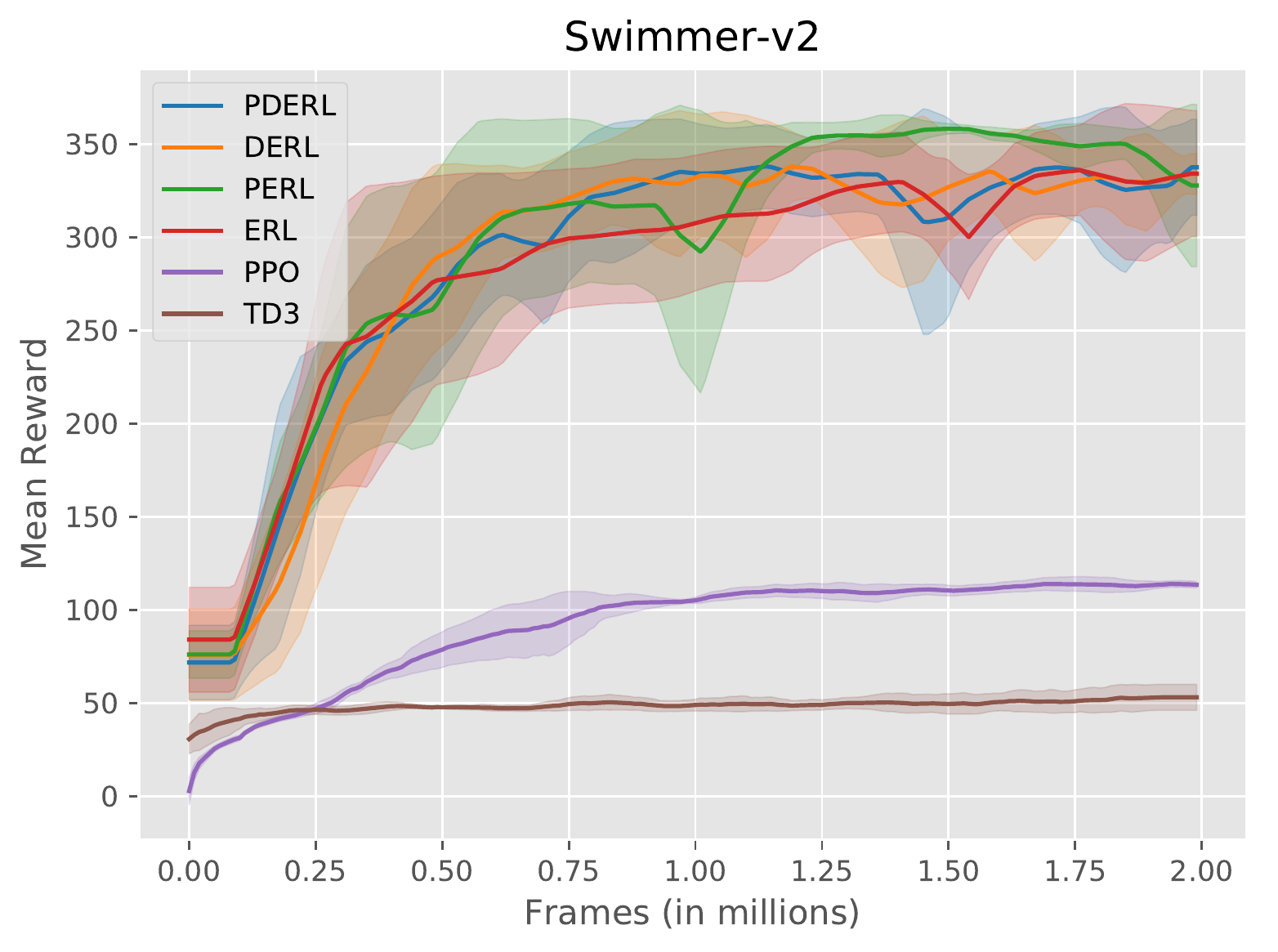}
         \caption{Swimmer-v2}
         \label{fig:results_swimmer}
    \end{subfigure}
    ~
    \begin{subfigure}[t]{0.30\textwidth}
         \centering
         \includegraphics[width=\columnwidth]{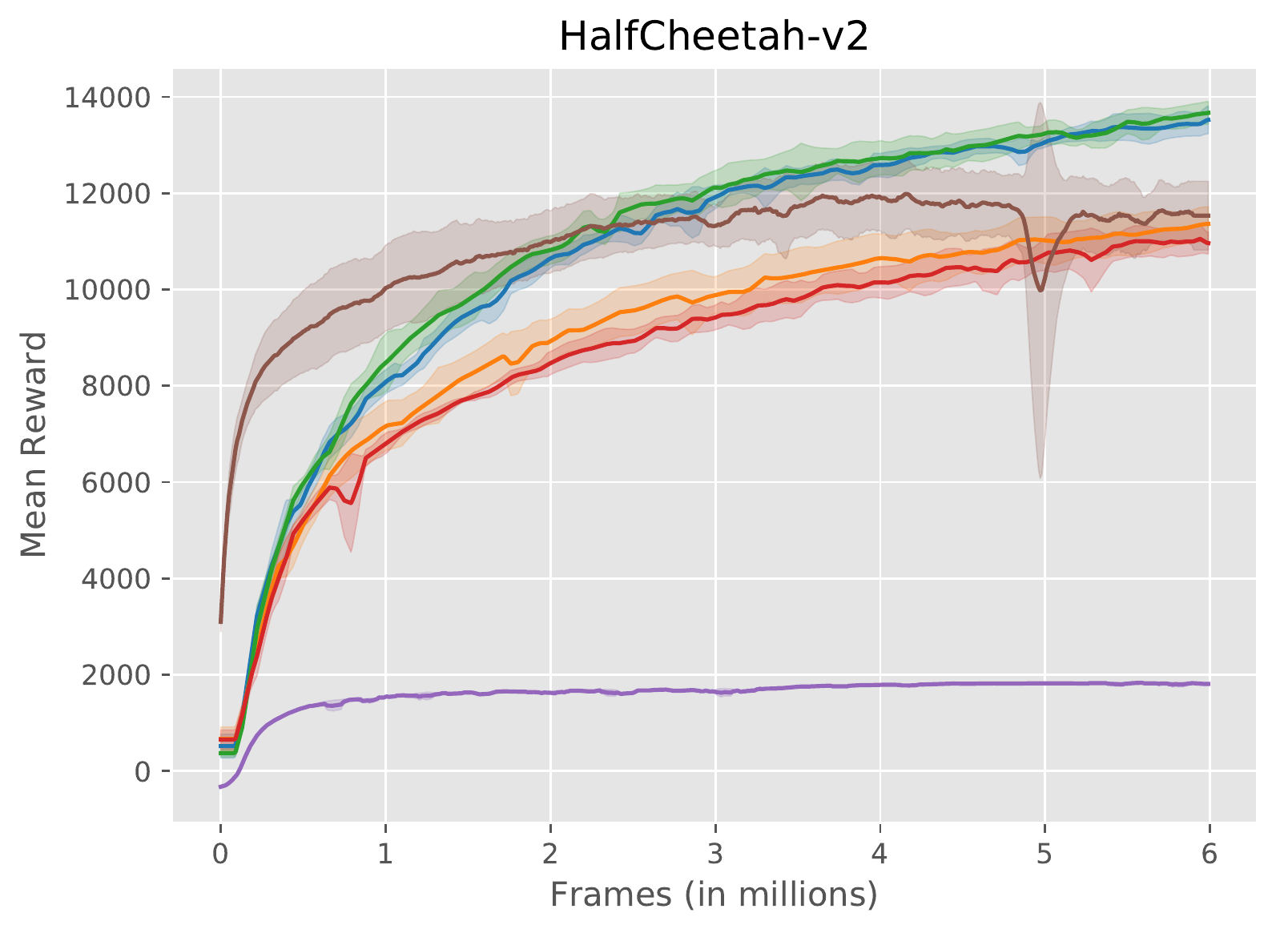}
         \caption{HalfCheetah-v2}
         \label{fig:results_cheetah}
    \end{subfigure}
    ~ 
    \begin{subfigure}[t]{0.30\textwidth}
        \centering
        \includegraphics[width=\columnwidth]{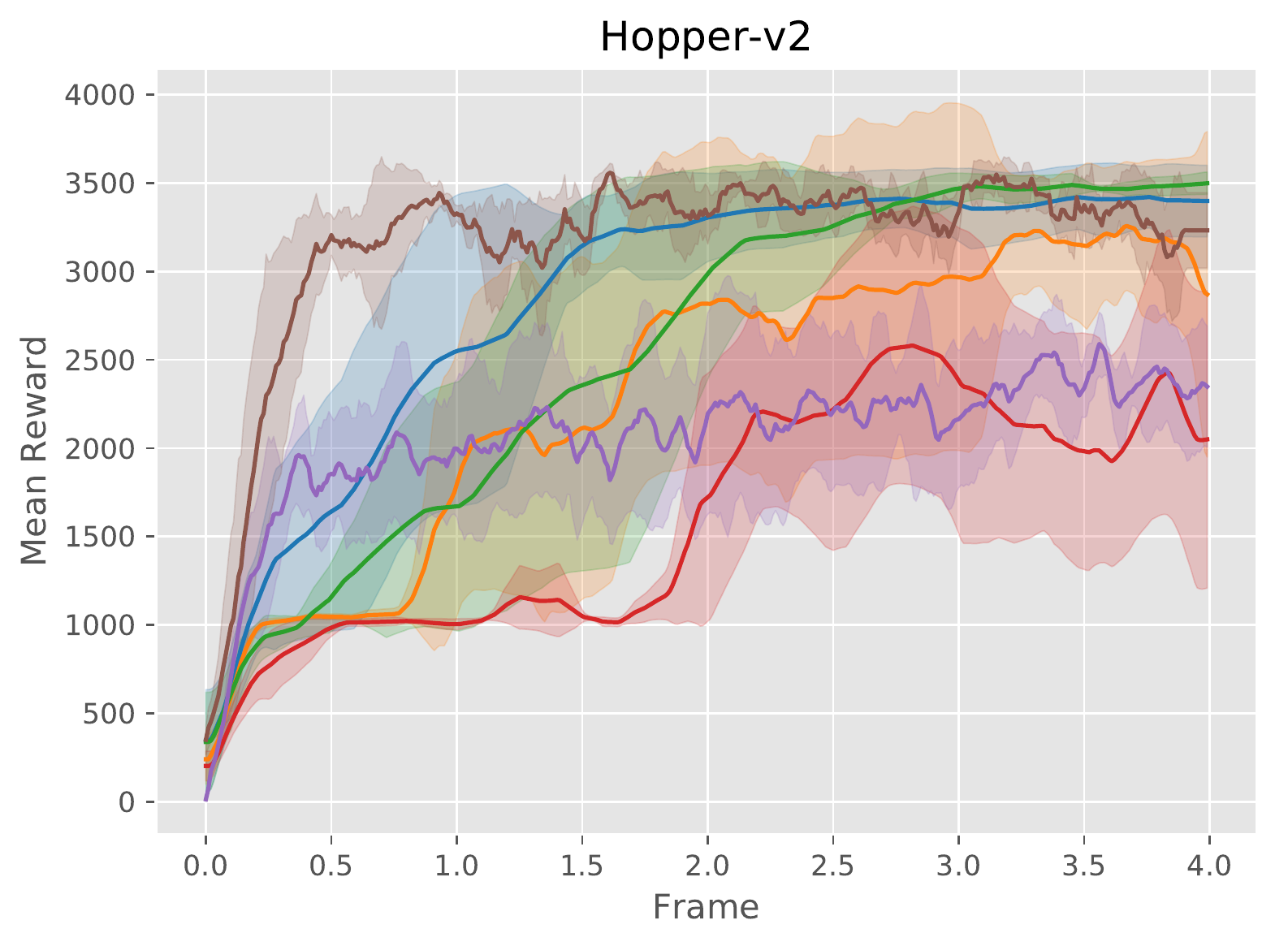}
        \caption{Hopper-v2}
        \label{fig:results_hopper}
    \end{subfigure}
    ~ 
     \begin{subfigure}[t]{0.30\textwidth}
       \centering
        \includegraphics[width=\columnwidth]{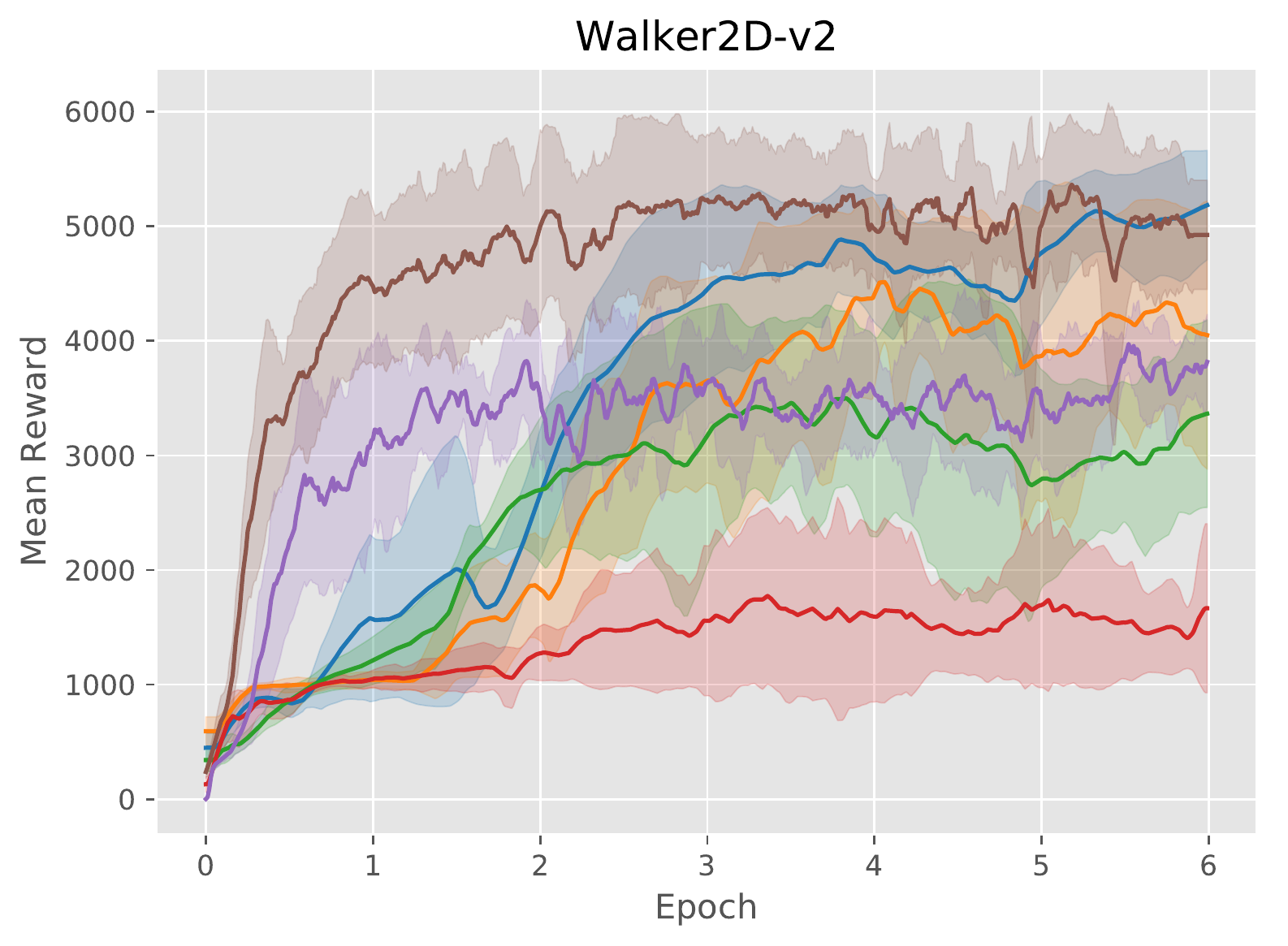}
        \caption{Walker2d-v2}
        \label{fig:results_walker}
    \end{subfigure}
    ~
   \begin{subfigure}[t]{0.30\textwidth}
       \centering
        \includegraphics[width=\columnwidth]{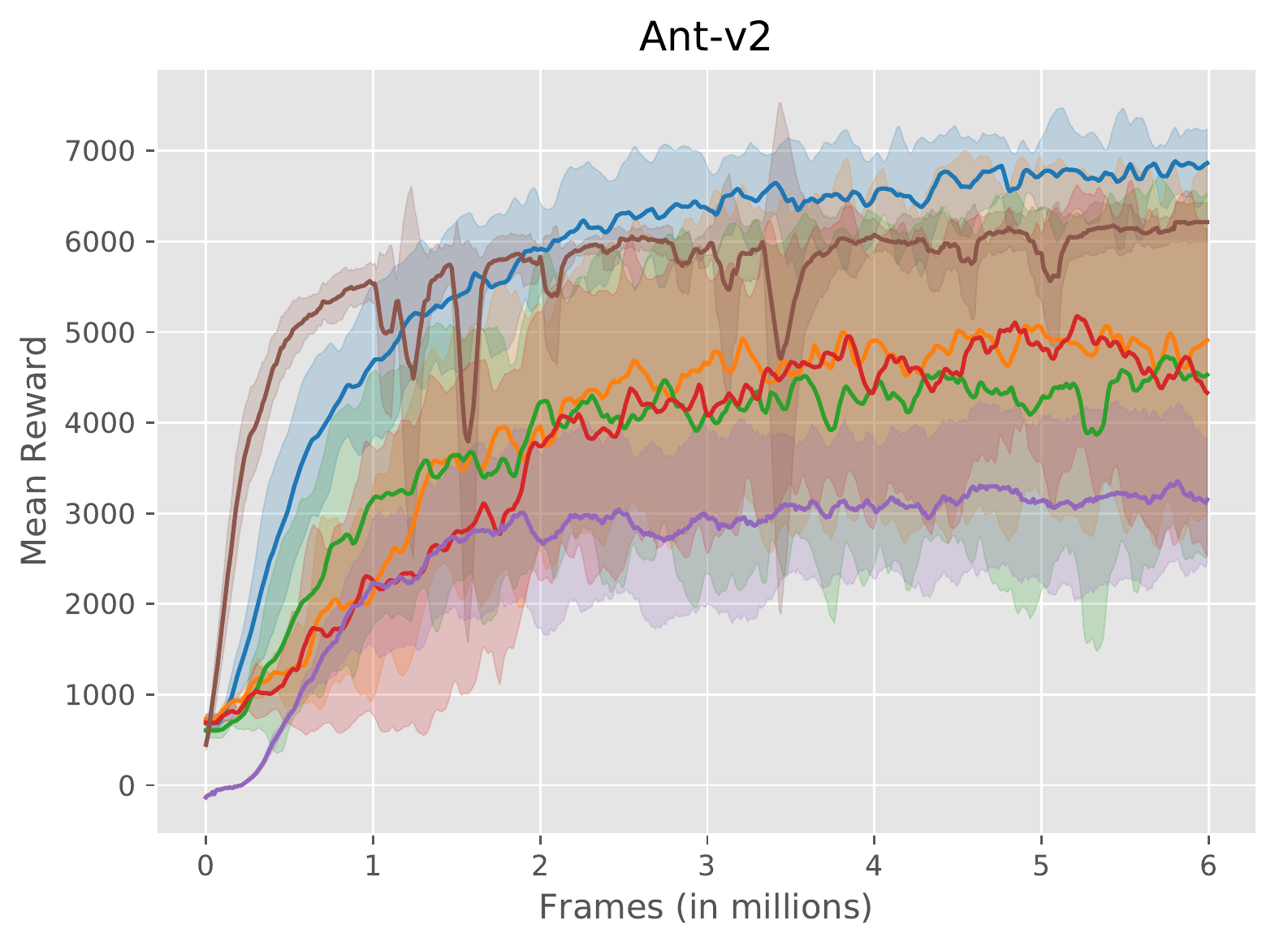}
        \caption{Ant-v2}
        \label{fig:results_ant}
    \end{subfigure}
    \caption[Mean reward evaluation]{The mean reward obtained on Swimmer (a), HalfCheetah (b), Hopper (c), Walker2d (d) and Ant (e). PDERL outperform ERL, PPO and TD3 on all the environments.}
    \label{fig:results}
\end{figure*}

This section evaluates the mean reward obtained by the newly proposed methods as a function of the number of environment frames experienced. The results are reported across five random seeds. Figure \ref{fig:results} shows the mean reward and the standard deviation obtained by all algorithms on five MuJoCo \citep{MuJoCo} environments.

While PERL and DERL bring improvements across multiple environments, they do not perform well across all of them. PERL is effective in stable environments like HalfCheetah and Hopper, where the total reward has low variance over multiple rollouts. At the same time, DERL is more useful in unstable environments like Walker2d and Ant since it drives the population towards regions with higher $Q$ values. In contrast, PDERL performs consistently well across all the settings, demonstrating that the newly introduced operators are complementary. PDERL significantly outperforms ERL and PPO across all environments and, despite being generally less sample efficient than TD3, it catches up eventually. Ultimately, PDERL significantly outperforms TD3 on Swimmer, HalfCheetah and Ant, and marginally on Hopper and Walker2d.


Table \ref{tab:results} reports the final reward statistics for all the tested models and environments. Side by side videos of ERL and PDERL running on simulated robots can be found at \url{https://youtu.be/7OGDom1y2YM}. The following subsections take a closer look at the newly introduced operators and offer a justification for the improvements achieved by PDERL.

\begin{table}[ht]
    \centering
    \begin{adjustbox}{width=1.0\columnwidth}
    \small
    \begin{tabular}{ll|cccccc}
    \toprule
    \textbf{Environment} & \textbf{Metric}  & \textbf{TD3} & \textbf{PPO} & \textbf{ERL} & \textbf{PERL} & \textbf{DERL} & \textbf{PDERL} \\ \midrule
    \multirow{3}{*}{Swimmer} & Mean & 53 & 113 & 334 & 327 & 333 & \textbf{337} \\
     & Std. & 26 & 3 & 20 & 26 & 6 & \textbf{12} \\
     & Median & 51 & 114 & 346 & 354 & 338 & \textbf{348} \\
    \midrule
    \multirow{3}{*}{HalfCheetah} & Mean & 11534 & 1810 & 10963 & \textbf{13668} & 11362 & 13522 \\
     & Std. & 713 & 28 & 225 & \textbf{236} & 358 & 287 \\
     & Median & 11334 & 1810 & 11025 & \textbf{13625} & 11609 & 13553 \\
    \midrule
    \multirow{3}{*}{Hopper} & Mean & 3231 & 2348 & 2049 & \textbf{3497} & 2869 & 3397 \\
     & Std. & 213 & 342 & 841 & \textbf{63} & 920 & 202 \\
     & Median & 3282 & 2484 & 1807 & \textbf{3501} & 3446 & 3400 \\
    \midrule
    \multirow{3}{*}{Walker2D} & Mean & 4925 & 3816 & 1666 & 3364 & 4050 & \textbf{5184} \\
     & Std. & 476 & 413 & 737 & 818 & 1170 & \textbf{477} \\
     & Median & 5190 & 3636 & 1384 & 3804 & 4491 & \textbf{5333} \\
    \midrule
     \multirow{3}{*}{Ant} & Mean & 6212 & 3151 & 4330 & 4528 & 4911 & \textbf{6845} \\
     & Std. & 216 & 686 & 1806 & 2003 & 1920 & \textbf{407} \\
     & Median & 6121 & 3337 & 5164 & 3331 & 5693 & \textbf{6948} \\
    \bottomrule
    \end{tabular}
    \end{adjustbox}
    \caption[Final performance]{Final performance in all environments. The result with the highest mean is shown in bold. PERL marginally outperforms PDERL on two environments, but PDERL consistently performs well across all environments.}
    \label{tab:results}
\end{table}

\begin{figure}[!ht]
    \centering
    \includegraphics[width=1.0\columnwidth]{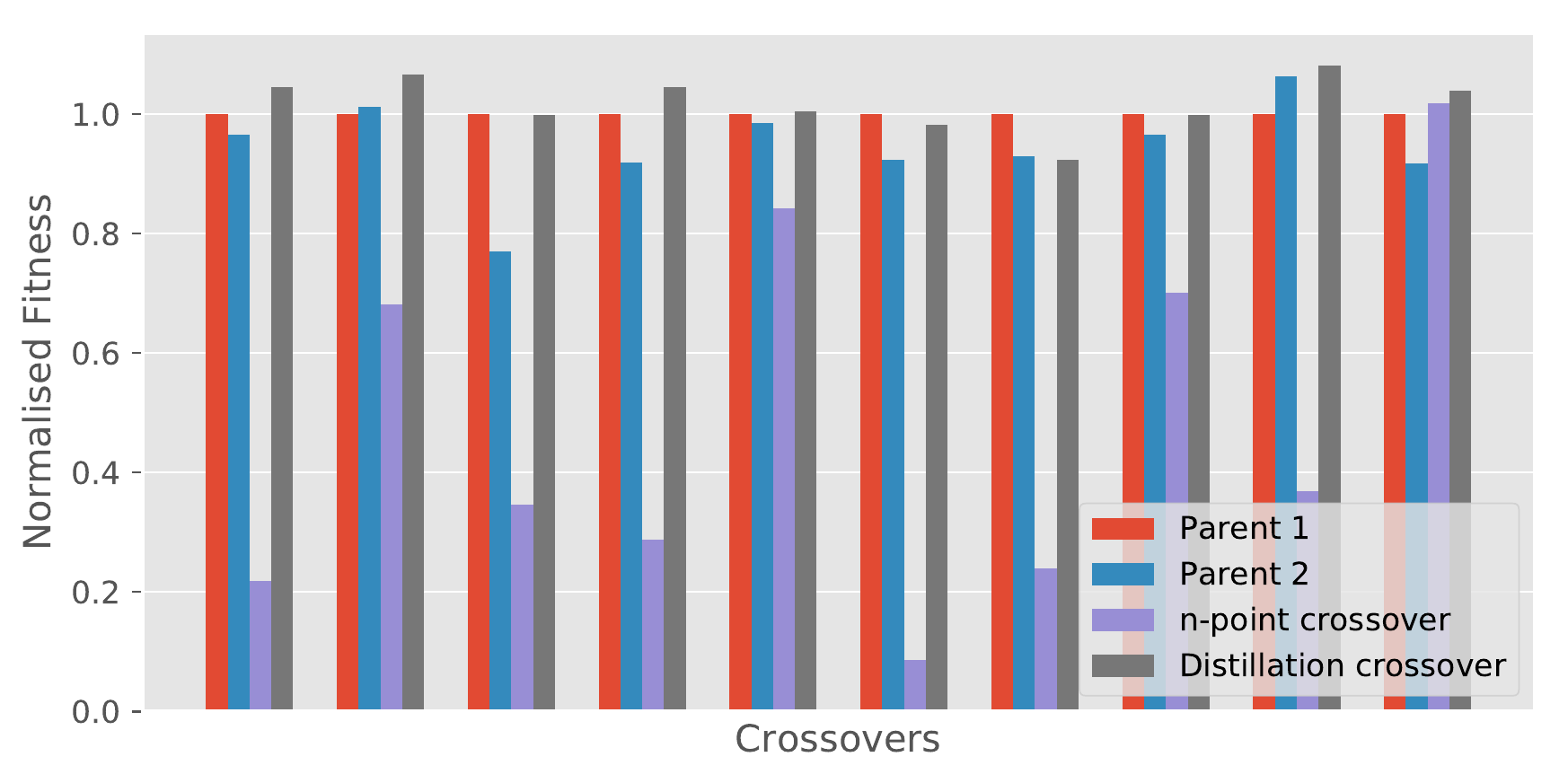}
    \caption[Crossover performance]{Normalised crossover performance on the Ant environment. The distillation crossover achieves higher fitness than the $n$-point crossover. Fitness is relative to Parent 1 in each group.}
    \label{fig:crossover_fit}
\end{figure}

\subsection{Crossover evaluation}

A good indicator for the quality of a crossover operator is the fitness of the offspring compared to that of the parents. Figure \ref{fig:crossover_fit} plots this metric for ten randomly chosen pairs of parents in the Ant environment. Each group of bars gives the fitness of the two parents and the policies obtained by the two types of crossovers. All these values are normalised by the fitness of the first parent. The performance of the child obtained via an $n$-point crossover regularly falls below $40\%$ the fitness of the best parent. At the same time, the fitness of the policies obtained by distillation is generally at least as good as that of the parents. 

\begin{figure}[!ht]
    \centering
    \includegraphics[width=1.0\columnwidth]{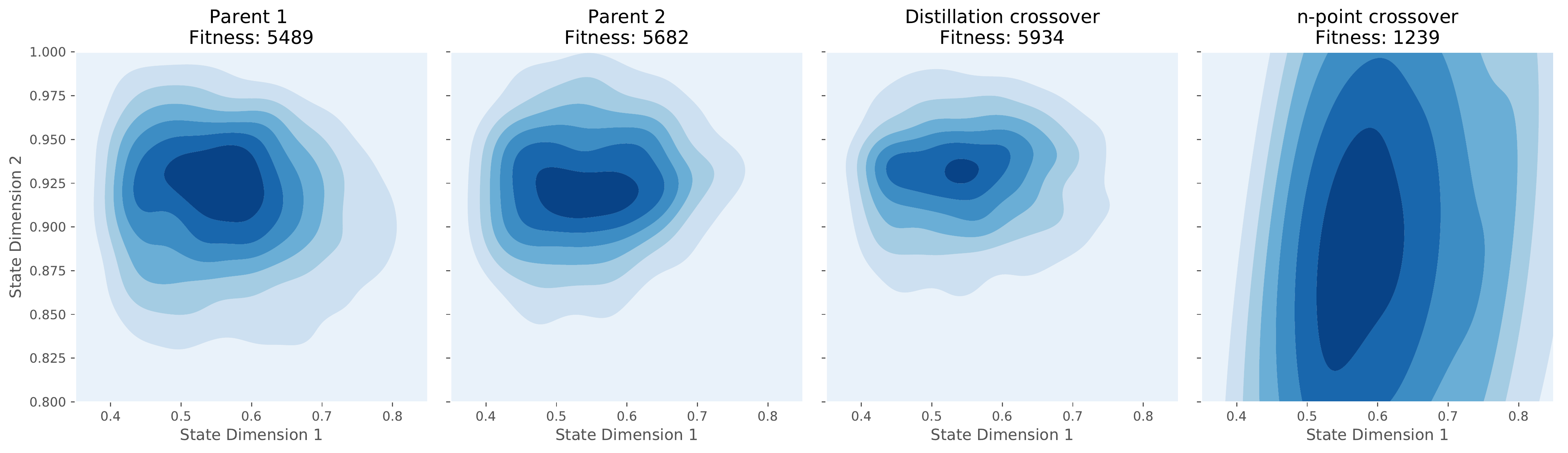}
    \caption[Crossover state visitation distribution]{This figure shows the state visitation distributions for the distillation crossover and the $n$-point crossover. Unlike the $n$-point crossover, the distillation crossover produces policies that selectively merge the behaviour of the parents.}
    \label{fig:crossover_state_1}
\end{figure}

The state visitation distributions of the parents and children offer a clearer picture of the two operators. Figure \ref{fig:crossover_state_1} shows these distributions for a sample crossover in the Ant environment. The $n$-point crossover produces a behaviour that diverges from that of the parents. In contrast, the $Q$-filtered distillation crossover generates a policy whose behaviour contains the best traits of the parent behaviours. The new operator implicitly drives each new generation in the population towards regions with higher $Q$ values.

\subsection{Mutation evaluation}

\begin{figure}[ht]
    \centering
    \includegraphics[width=1.0\columnwidth]{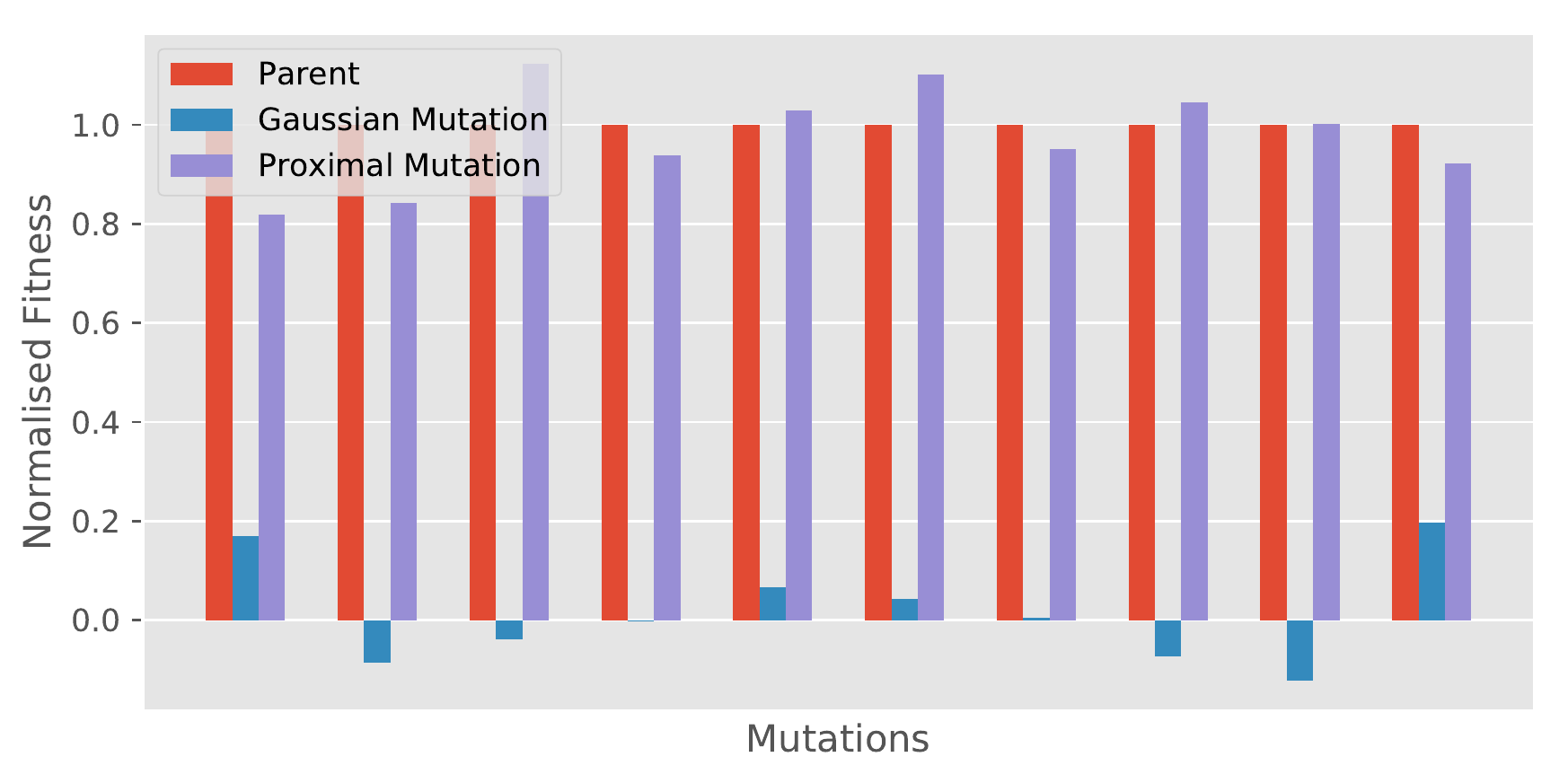}
    \caption[Mutation performance]{Normalised mutation performance on the Ant environment. The proximal mutations obtain significantly higher fitness than the Gaussian mutations. Fitness is relative to the Parent in each group.}
    \label{fig:mutation_fit}
\end{figure}

Figure \ref{fig:mutation_fit} shows the fitness of the children obtained by the two types of mutation for ten randomly selected parents on the Ant environment. Most Gaussian mutations produce child policies with fitness that is either negative or close to zero. At the same time, the proximal mutations create individuals that often surpass the fitness of the parents. 

\begin{figure}[!ht]
    \centering
    \includegraphics[width=1.0\columnwidth]{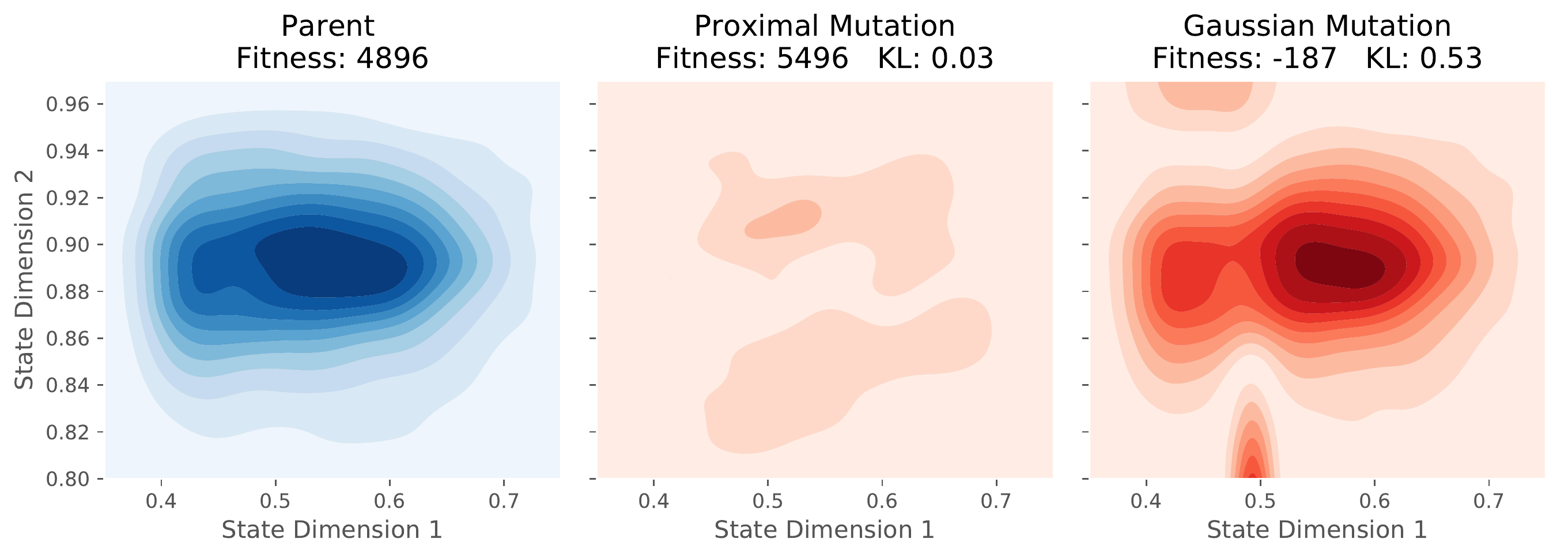}
    \caption[Mutation state visitation distribution]{As before, the blue contours represent the state visitation distribution of the parent, whereas the red ones represent the difference. The child obtained by proximal mutation inherits the behaviour of the parent to a large degree and obtains a $600$ fitness boost. The behaviour obtained by Gaussian mutation is entirely different from that of the parent. The KL divergence between the parent and child distributions (0.03 and 0.53) quantitatively confirm this.}
    \label{fig:mutation_states_1}
\end{figure}

As in the previous section, the analysis of the state visitation distribution of the policies reveals the destructive behaviour of the Gaussian mutations. The contours of these distributions for a sample mutation are given in Figure \ref{fig:mutation_states_1}. The policy mutated by additive Gaussian noise completely diverges from the behaviour of the parent. This sudden change in behaviour causes catastrophic forgetting, and the new offspring falls in performance to a total reward of $-187$. In contrast, the proximal mutation generates only a subtle change in the state visitation distribution. The offspring thus obtained inherits to a great extent the behaviour of the parent, and achieves a significantly higher total reward of $5496$.

\section{Related work}

This paper is part of an emerging direction of research attempting to merge Evolutionary Algorithms and Deep Reinforcement Learning: \citet{Khadka2018EvolutionGuidedPG}, \citet{Pourchot2019CEMRLCE}, \citet{Gangwani2018PolicyOB}, \citet{pmlr-v97-khadka19a}.


Most closely related are the papers of \citet{Lehman2018SafeMF} and \citet{Gangwani2018PolicyOB}. Both of these works address the destructive behaviours of classic variation operators. \citet{Lehman2018SafeMF} focus exclusively on safe mutations, and one of their proposed operators is directly employed in the proximal mutations. However, their paper is lacking a treatment of crossovers and the integration with learning explored here. The methods of \citet{Gangwani2018PolicyOB} are focused exclusively on safe operators for stochastic policies, while the methods proposed in this work can be applied to stochastic and deterministic policies alike. The closest aspect of their work is that they also introduce a crossover operator with the goal of merging the behaviour of two agents. Their solution reduces the problem to the traditional single parent distillation problem using a maximum-likelihood approach to combine the behaviours of the two parents. They also propose a mutation operator based on gradient ascent using policy gradient methods. However, this deprives their method of the benefits of derivative-free optimisation such as the robustness to local optima. 

\section{Discussion}

The ERL framework demonstrates that genetic algorithms can be scaled to DNNs when combined with learning methods. In this paper we have proposed the PDERL extension and shown that performance is further improved with a hierarchical integration of learning and evolution. While maintaining a bi-directional flow of information between the population and RL agent, our method also uses learning within the genetic operators which, unlike traditional implementations, produce the desired functionality when applied to directly encoded DNNs. Finally, we show that PDERL outperforms ERL, PPO and TD3 in all tested environments.

Many exciting directions for future research remain, as discussed in the text. An immediate extension would be to develop a distributed version able to exploit larger and more diverse populations. Better management of the inherited genetic memories may yield efficiency gains by prioritising key experiences. Lastly, we note the potential for using learning algorithms at the level of selection operators.

\bibliography{ref}
\bibliographystyle{aaai}


\newpage
\section{Supplementary Material}

\subsection{Behaviour and mutation magnitude}

\begin{figure}[!ht]
    \centering
    \includegraphics[width=\columnwidth]{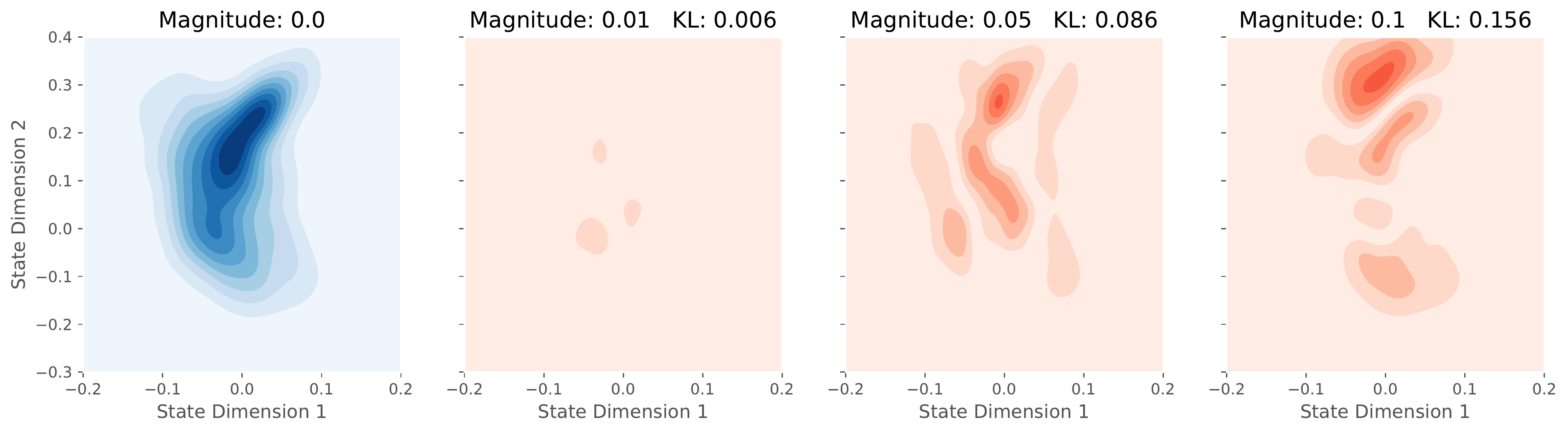}
    \caption[Mutation magnitude and behaviour]{This figure shows how the state visitation distributions of the child policies obtained by proximal mutation smoothly depend on the mutation magnitude. The blue contour corresponds to the parent's state visitation distribution. The red contours show how the differences from the parent distribution increase as the mutation magnitude becomes higher. The KL divergence between the parent and child distributions, which strictly increases with the mutation magnitude, also confirm this behaviour. }
    \label{fig:mut_abl}
\end{figure}

Another benefit of the proximal mutations is that the size of the change in the behaviour can be directly adjusted by tuning the magnitude of the mutations. Figure \ref{fig:mut_abl} shows how increasing the mutation magnitude gradually induces a more significant change in the behaviour of the child policy. This is not the case for the Gaussian mutations where a given mutation size can be unpredictably large and non-linearly related to the behaviour changes. 

\subsection{Parent selection mechanism}

The previous experiments used the greedy parent selection mechanism for choosing the policies involved in the crossovers. This section offers a comparative view between this greedy selection and the newly proposed distance-based selection. 

\begin{figure}[!ht]
    \centering
        \begin{subfigure}[t]{0.9\columnwidth}
         \centering
         \includegraphics[width=\columnwidth]{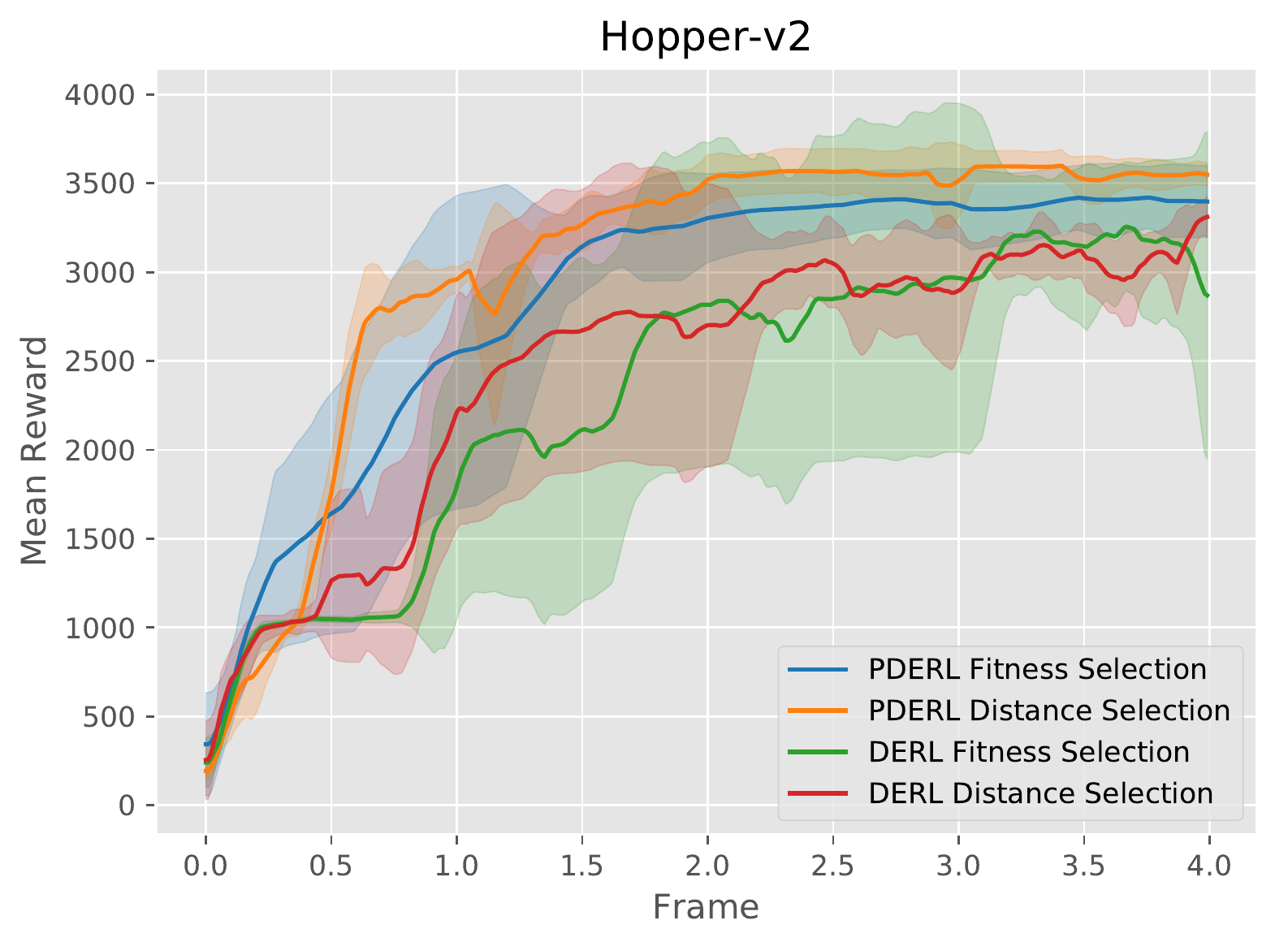}
         \caption{Hopper-v2}
         \label{fig:parent_selection_hopper}
    \end{subfigure}
    ~
    \begin{subfigure}[t]{0.9\columnwidth}
         \centering
         \includegraphics[width=\columnwidth]{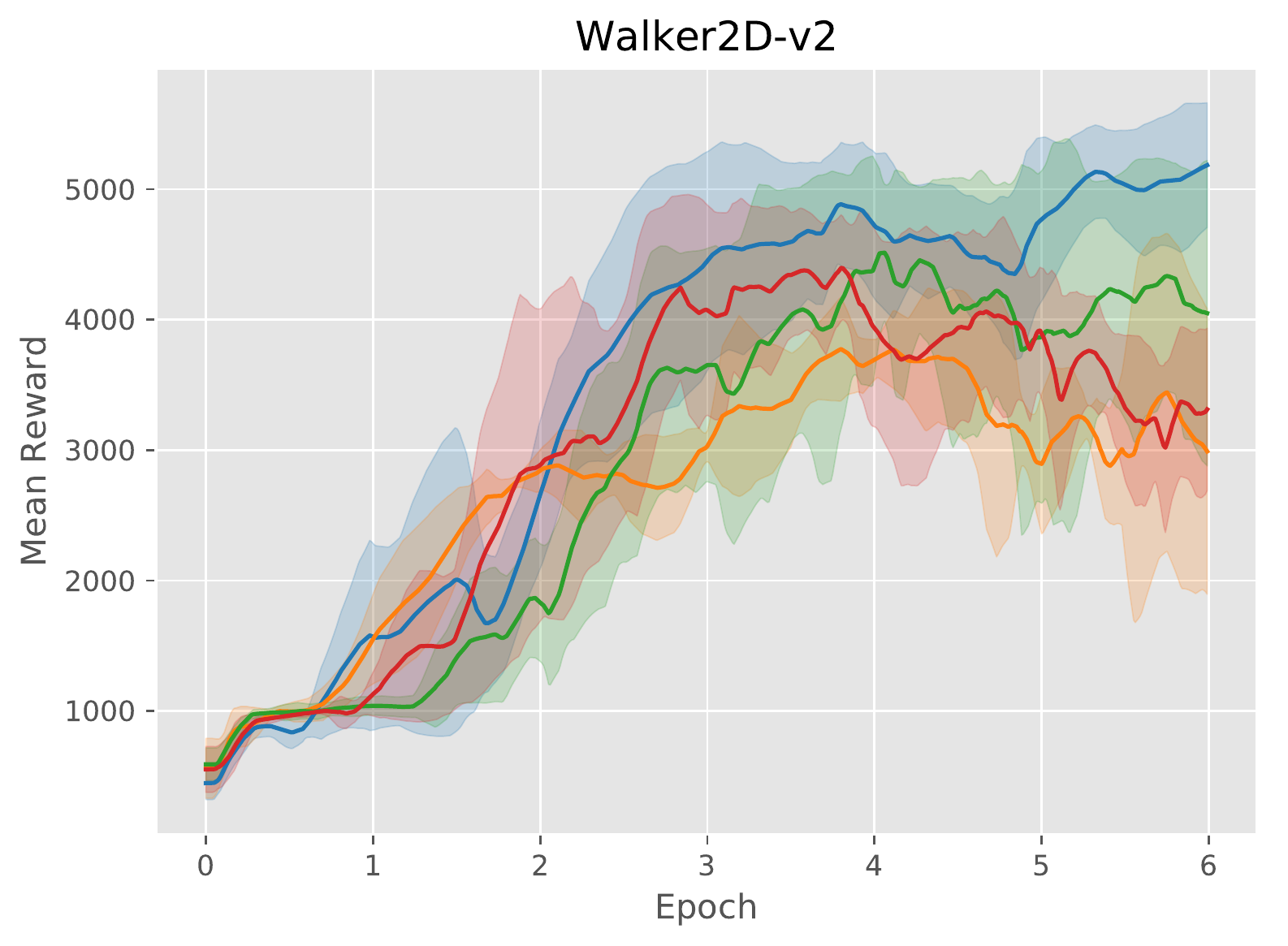}
         \caption{Walker2d-v2}
         \label{fig:parent_selection_walker}
    \end{subfigure}
    \caption[Comparison of parent selection mechanisms]{A comparison between the greedy (fitness-based) parent selection mechanism and the distance-based parent selection. The distance-based selection works better in stable environments like Hopper, but worse in environments with high variance across different episodes like Walker2d.}
    \label{fig:parent_selection}
\end{figure}

The mechanisms are compared in Figure \ref{fig:parent_selection} on Hopper and Walker2d. On a relatively unstable environment like Walker2d, the distance-based DERL and PDERL perform significantly worse than their fitness-based equivalents. However, on Hopper, which is more stable than Walker2d, the distance-based PDERL surpasses the fitness-based one, while also having very low variance.

The fact that the distance-based selection performs better in the early stages of training for both environments supports the idea of using a convex combination of the two selections.

\subsection{Interactions between learning and evolution}
\label{sec:interactions}

\begin{figure*}[!ht]
    \centering
        \begin{subfigure}[t]{0.3\textwidth}
         \centering
         \includegraphics[width=\textwidth]{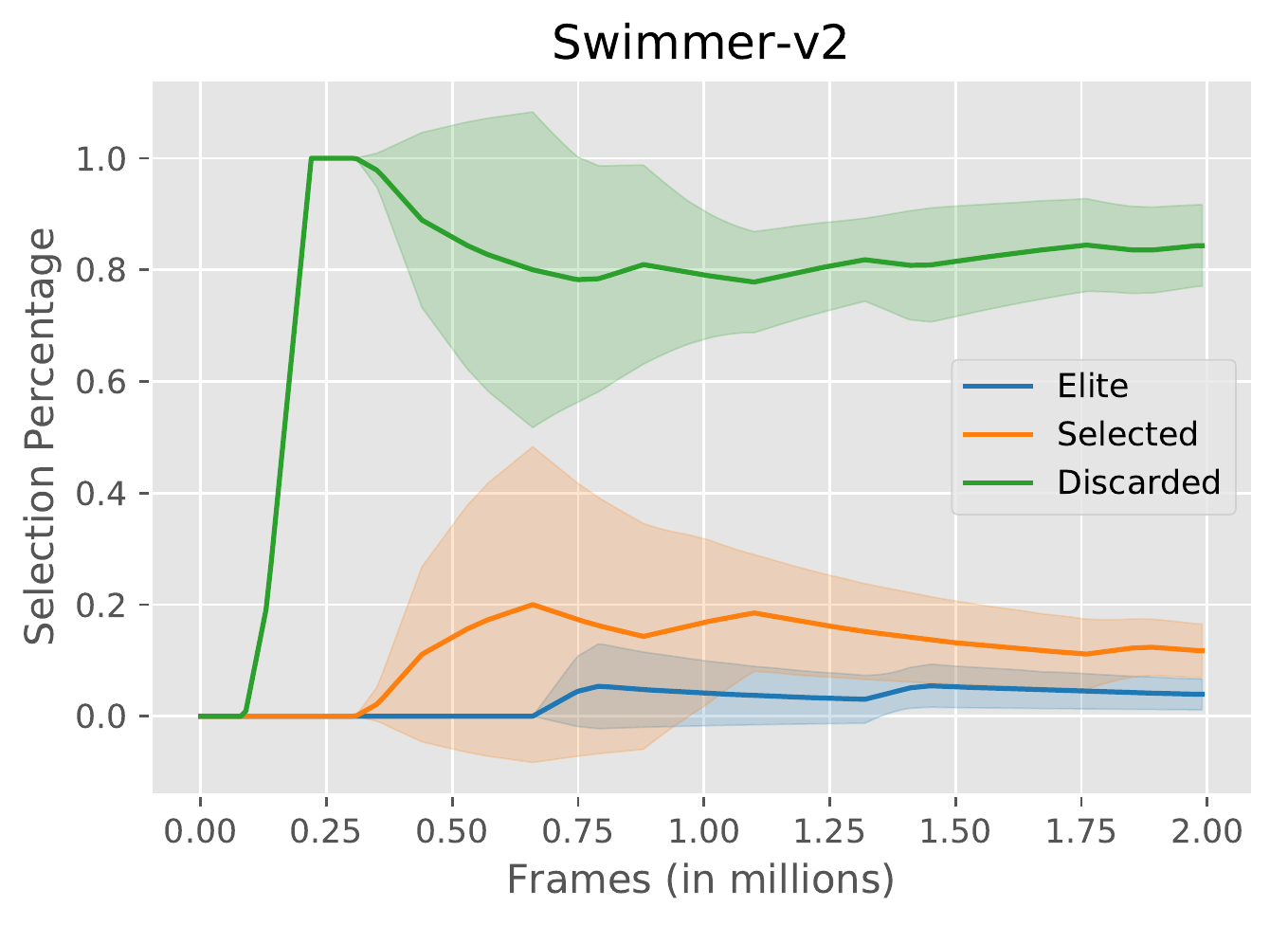}
         \caption{Swimmer-v2}
         \label{fig:swimmer_select}
    \end{subfigure}
    ~
    \begin{subfigure}[t]{0.3\textwidth}
         \centering
         \includegraphics[width=\textwidth]{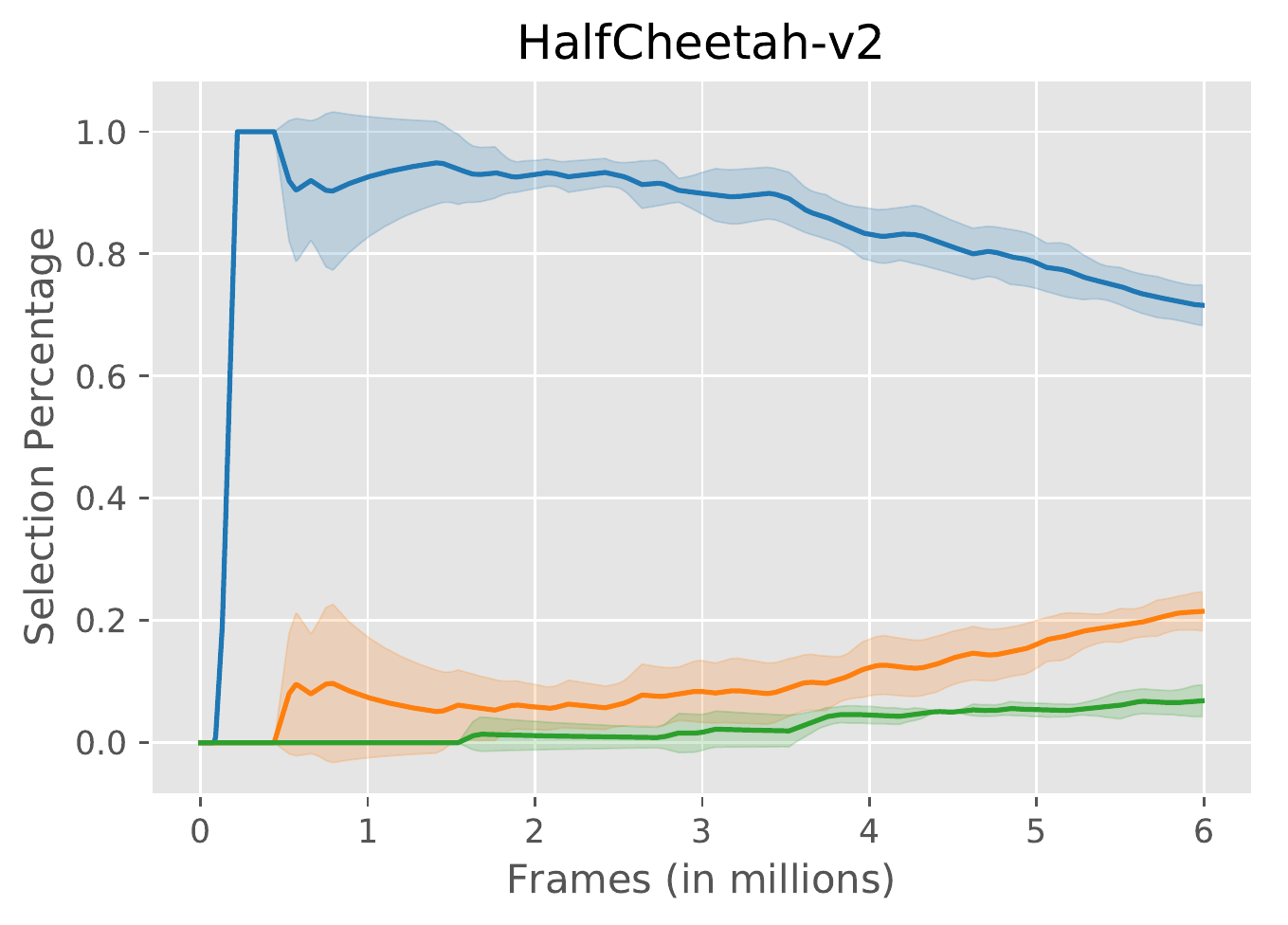}
         \caption{HalfCheetah-v2}
         \label{fig:cheetah_select}
    \end{subfigure}
    ~
    \begin{subfigure}[t]{0.3\textwidth}
         \centering
         \includegraphics[width=\textwidth]{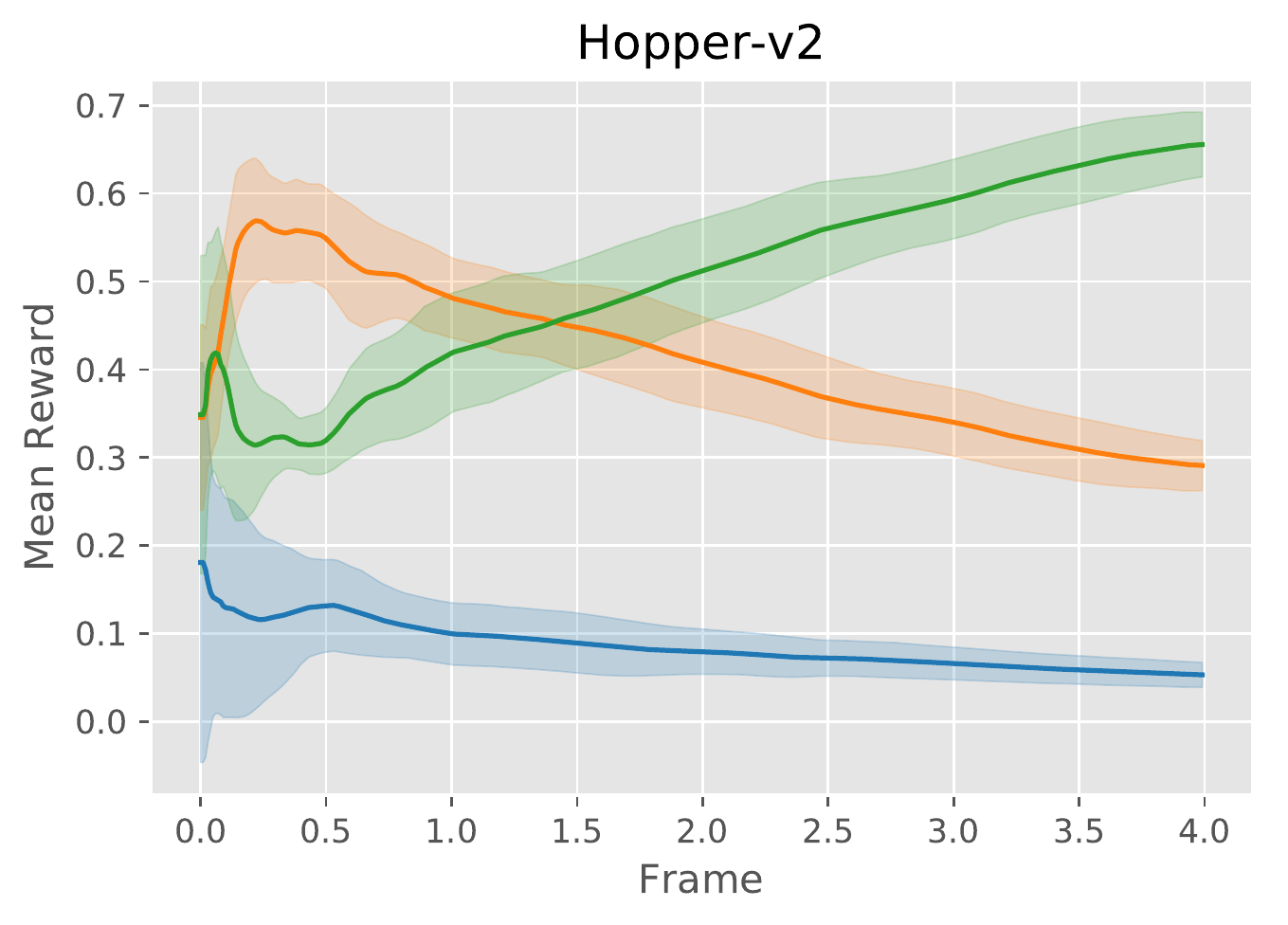}
         \caption{Hopper-v2}
         \label{fig:hopper_select}
    \end{subfigure}
    ~
    \begin{subfigure}[t]{0.3\textwidth}
         \centering
         \includegraphics[width=\textwidth]{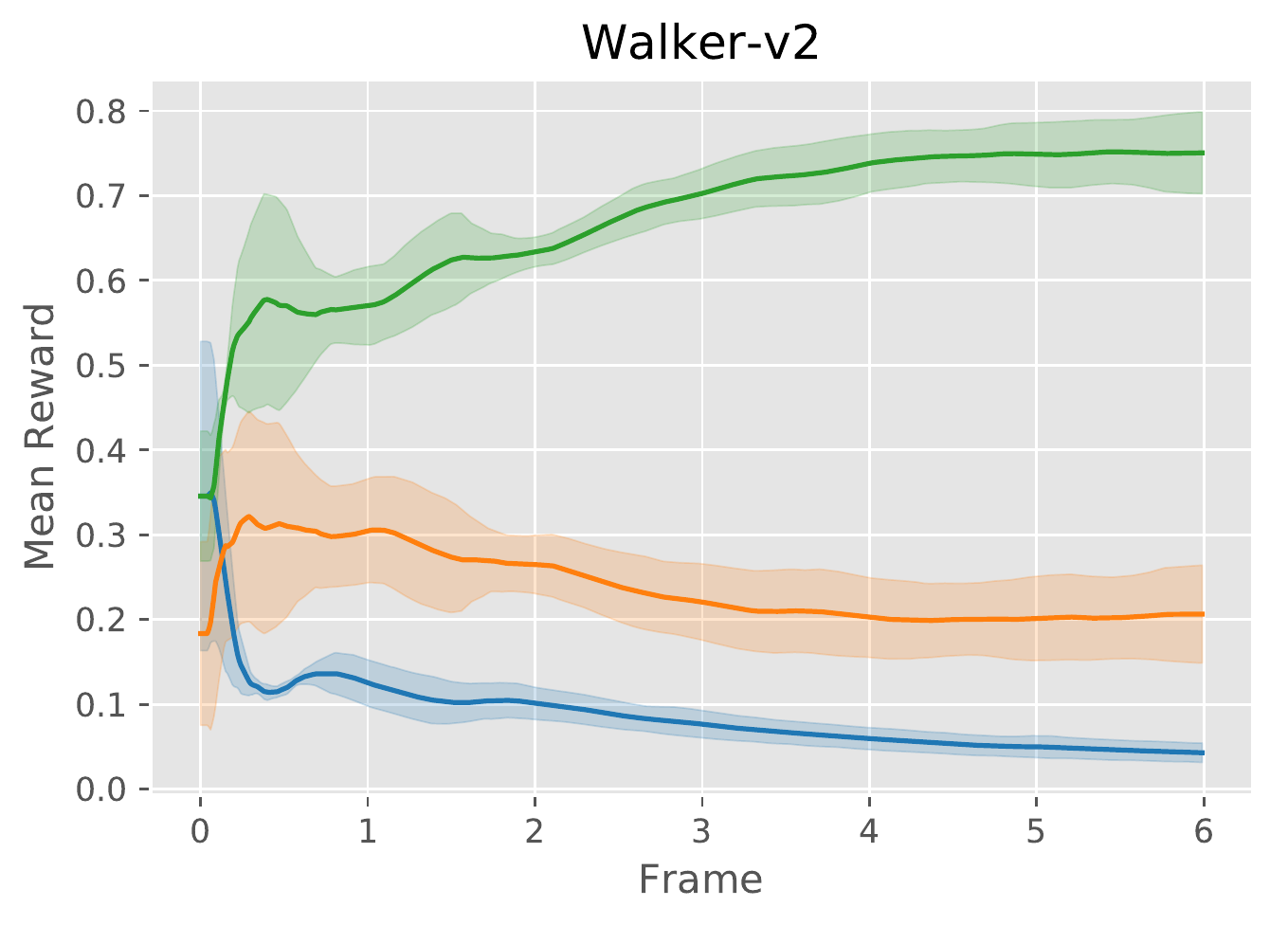}
         \caption{Walker2d-v2}
         \label{fig:walker_select}
    \end{subfigure}
    ~
    \begin{subfigure}[t]{0.3\textwidth}
      \centering
        \includegraphics[width=\textwidth]{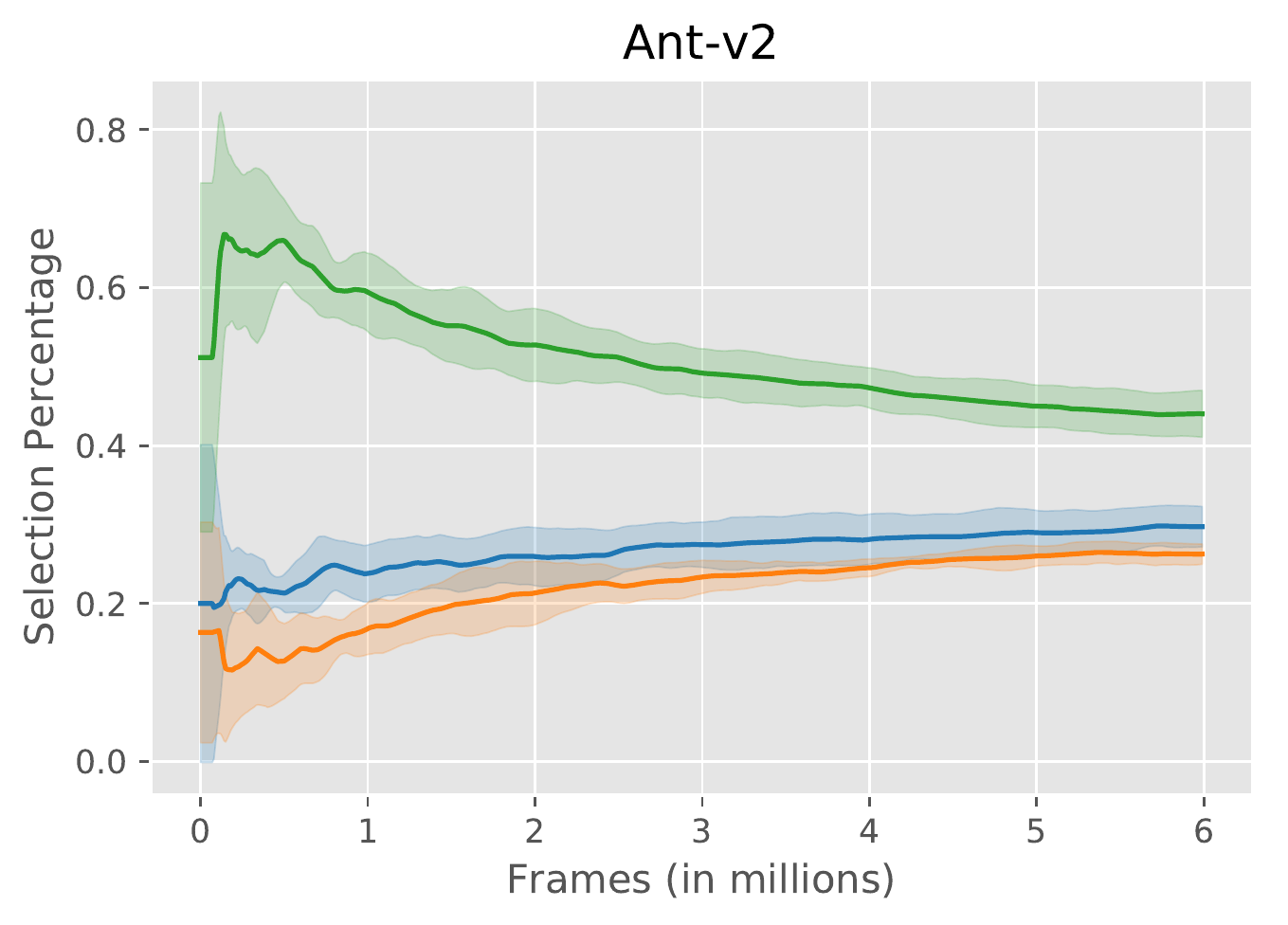}
        \caption{Ant-v2}
        \label{fig:ant_select}
    \end{subfigure}
    \caption[Interactions between learning and evolution]{Selection rates for the RL agent over the course of training on Swimmer (a), HalfCheetah (b), Hopper (c), Walker2d (d) and Ant (e). PDERL produces on each environment a unique and rich interaction pattern between evolution and learning.}
    \vskip -0.2in

    \label{fig:selection}
\end{figure*}

One of the motifs of this work is the interaction between learning and evolution. In PDERL, this can be quantitatively analysed by looking at the number of times the RL agent was selected, discarded or became an elite in the population. Figure \ref{fig:selection} shows how these numbers evolve during training and reveals that each environment has its own unique underlying interaction pattern.

\begin{table}[!ht]
    \newcolumntype{C}{ @{}>{${}}c<{{}$}@{} }
    \centering
    \begin{adjustbox}{width=1.0\columnwidth}
    \small
    \begin{tabular}{ll *3{rCl}}
        \toprule
        \textbf{Environment} & \textbf{Algorithm} & \multicolumn{3}{c}{\textbf{Elite}} & \multicolumn{3}{c}{\textbf{Selected}} & \multicolumn{3}{c}{\textbf{Discarded}} \\ 
        \midrule
        \multirow{2}{*}{Swimmer}        & ERL   & 4.0  & \pm & 2.8      & 20.3 & \pm & 18.1  & 76.0 & \pm & 20.4 \\
                                        & PDERL & 3.9  & \pm & 2.7      & 11.7 & \pm & 4.7   & 84.3 & \pm & 7.3 \\ 
        
        \multirow{2}{*}{HalfCheetah}    & ERL   & 83.8 & \pm & 9.3      & 14.3 & \pm & 9.1   & 2.3  & \pm & 2.5 \\
                                        & PDERL & 71.6 & \pm & 3.3      & 21.5 & \pm & 3.1   & 6.8  & \pm & 2.6 \\ 
        
        \multirow{2}{*}{Hopper}         & ERL   & 28.7 & \pm & 8.5      & 33.7 & \pm & 4.1   & 37.7 & \pm & 4.5 \\
                                        & PDERL & 5.3  & \pm & 1.4      & 29.0 & \pm & 2.7   & 65.6 & \pm & 3.6 \\ 
        
        \multirow{2}{*}{Walker2D}       & ERL   & 38.5 & \pm & 1.5      & 39.0 & \pm & 1.9   & 22.5 & \pm & 0.5 \\ 
                                        & PDERL & 4.3  & \pm & 1.5      & 20.6 & \pm & 5.7   & 75.0 & \pm & 4.8 \\ 
        \multirow{2}{*}{Ant}            & ERL   & 66.7 & \pm & 1.7      & 15.0 & \pm & 1.4   & 18.0 & \pm & 0.8 \\ 
                                        & PDERL & 29.7 & \pm & 2.5      & 26.2 & \pm & 1.2   & 44.0 & \pm & 2.9 \\  
        \bottomrule
    \end{tabular}
    \end{adjustbox}
    \caption[PDERL final selection rates]{Cumulative selection rates for the RL agent for ERL and PDERL.}
    \label{tab:selection}
\end{table}

\textbf{Swimmer}. Swimmer is an environment where the genetic population drives the progress of the agent almost entirely. The RL agent rarely becomes good enough to be selected or become an elite.

\textbf{HalfCheetah}. HalfCheetah is at the other end of the spectrum from Swimmer. In this environment, the RL agent becomes an elite after over $90\%$ of the synchronisations early in training. As the population reaches the high-reward areas, genetic evolution becomes slightly more important for discovering better policies.

\textbf{Hopper}. On Hopper, the RL agent drives the population in the first stages of training, but the population obtains superior total rewards beyond 1.5 million frames. Therefore, evolution has a greater contribution to the late stages of training.

\textbf{Walker2d}. In this environment, the dynamics between learning and evolution are very stable. The selection rates do not change significantly during the training process. This is surprising, given the instability of Walker2d. Overall, evolution has a much higher contribution than learning. 

\textbf{Ant}. Unlike in the other environments, in the Ant environment, the interactions between learning and evolution are more balanced. The curves converge towards a $40\%$ probability of being discarded, and a $60\%$ probability of becoming an elite or being selected.

Table \ref{tab:selection} shows the final mean selection rates and their standard deviations side by side with the ones reported by ERL. This comparison indicates how the dynamics between evolution and learning changed after adding the new crossovers and mutations. The general pattern that can be seen across all environments is that the probability that the RL agent becomes an elite decreases. That probability mass is mainly moved towards the cases when the RL agent is discarded. This comes as another confirmation that the newly introduced variation operators improve the performance of the population and the RL agent is much less often at the same level of fitness as the population policies.

\subsection{Hyperparameters}

Table \ref{tab:constant_hyperparams} includes the hyperparameters that where kept constant across all environments. Table \ref{tab:variable_hyperparams} specifies the parameters that vary with the task.

\begin{table}[ht]
    \centering
    \begin{tabular}{l c}
        \toprule
        \textbf{Hyperparameter} & \textbf{Value} \\ \midrule
        Population size $k$ & $10$ \\
        Target weight $\tau$ & $0.001$ \\
        RL Actor learning rate & $5e^{-5}$ \\
        RL Critic learning rate &  $5e^{-4}$ \\
        Genetic Actor learning rate & $1e{-3}$ \\
        Discount factor $\gamma$ & $0.99$ \\
        Replay buffer size & $1e^6$ \\
        Genetic memory size & $8000$ \\
        RL Agent batch size & $128$ \\
        Genetic Agent crossover batch size $N_{C}$ & $128$ \\
        Genetic Agent mutation batch size $N_M$ & $256$ \\
        Distillation crossover epochs & $12$ \\
        Mutation probability & $0.9$ \\ \bottomrule
        
    \end{tabular}
    \caption[Constant hyperparameters]{Hyperparameters constant across all environments.}
    \label{tab:constant_hyperparams}
\end{table}

\begin{table}[ht]
    \centering
    \begin{adjustbox}{width=1.0\columnwidth}
    \small
    \begin{tabular}{l ccccc}
    \toprule
    \textbf{Parameter} & \textbf{Swimmer} & \textbf{HalfCheetah} & \textbf{Hopper} & \textbf{Walker2D} & \textbf{Ant} \\
    \midrule
    Elite fraction $\psi$ & 0.1 & 0.1 & 0.2 & 0.2 & 0.2 \\
    Trials $\xi$ & 1 & 1 & 3 & 5 & 1 \\
    Sync. Period $\omega$ & 10 & 10 & 1 & 1 & 1 \\ 
    Mutation Mag. $\sigma$ & 0.1 & 0.1 & 0.1 & 0.1 & 0.01 \\ 
    \bottomrule
    \end{tabular} 
    \end{adjustbox}
    \caption[Variable hyperparameters]{Hyperparameters that vary across environments.}
    \label{tab:variable_hyperparams}
\end{table}

\end{document}